\documentclass[journal]{IEEEtran}
\usepackage{ulem}
\usepackage{float}
\usepackage{algorithm}
\usepackage{algpseudocode}
\usepackage[marginal]{footmisc}
\usepackage[numbers,square,comma,sort&compress]{natbib}

\usepackage{graphics} % for pdf, bitmapped graphics files
\usepackage{epsfig} % for postscript graphics files
\usepackage{mathptmx} % assumes new font selection scheme installed
\usepackage{times} % assumes new font selection scheme installed
\usepackage{indentfirst}
\usepackage{amsmath}
\usepackage{bbding}
\usepackage{amssymb}  % assumes amsmath package installed
\usepackage{booktabs}
\usepackage{url}
\usepackage{multirow}
\usepackage{bbding}
\usepackage{hyperref}
\usepackage{xcolor}
\usepackage{blindtext}

\usepackage{etoolbox}
\makeatletter
\patchcmd{\@makecaption}
  {\scshape}
  {}
  {}
  {}
\makeatletter
\patchcmd{\@makecaption}
  {\\}
  {.\ }
  {}
  {}
\makeatother

\ifCLASSINFOpdf
\else
\fi
\begin{document}
%
% paper title
% Titles are generally capitalized except for words such as a, an, and, as,
% at, but, by, for, in, nor, of, on, or, the, to and up, which are usually
% not capitalized unless they are the first or last word of the title.
% Linebreaks \\ can be used within to get better formatting as desired.
% Do not put math or special symbols in the title.
\title{Da Yu: Towards USV-Based Image Captioning for Waterway Surveillance and Scene Understanding}
%
%
% author names and IEEE memberships
% note positions of commas and nonbreaking spaces ( ~ ) LaTeX will not break
% a structure at a ~ so this keeps an author's name from being broken across
% two lines.
% use \thanks{} to gain access to the first footnote area
% a separate \thanks must be used for each paragraph as LaTeX2e's \thanks
% was not built to handle multiple paragraphs
%

\author{Runwei Guan,
        Ningwei Ouyang,
        Tianhao Xu,
        Shaofeng Liang,
        Wei Dai, 
        Yafeng Sun,
        Shang Gao,
        Songning Lai,
        Shanliang Yao,
        Xuming Hu,
        Ryan Wen Liu,
        Yutao Yue
        and Hui Xiong,~\IEEEmembership{~Fellow,~IEEE}% <-this % stops a space
\thanks{Runwei Guan, Tianhao Xu, Shang Gao, Songning Lai, Xuming Hu, Yutao Yue and Hui Xiong are with Hong Kong University of Science and Technology (Guangzhou). Hui Xiong is also with Hong Kong University of Science and Technology, Hong Kong SAR, China.}% <-this % stops a space
\thanks{Ningwei Ouyang and Wei Dai are with Xi'an Jiaotong-Liverpool University.}
\thanks{Shaofeng Liang is with China University of Petroleum (East China).}
\thanks{Yafeng Sun is with University of Science and Technology of China.}
\thanks{Shanliang Yao is with Yancheng Institute of Technology.}
\thanks{Ryan Wen Liu is with Wuhan University of Technology.}
\thanks{Corresponding author: xionghui@ust.hk}
}

% note the % following the last \IEEEmembership and also \thanks - 
% these prevent an unwanted space from occurring between the last author name
% and the end of the author line. i.e., if you had this:
% 
% \author{....lastname \thanks{...} \thanks{...} }
%                     ^------------^------------^----Do not want these spaces!
%
% a space would be appended to the last name and could cause every name on that
% line to be shifted left slightly. This is one of those "LaTeX things". For
% instance, "\textbf{A} \textbf{B}" will typeset as "A B" not "AB". To get
% "AB" then you have to do: "\textbf{A}\textbf{B}"
% \thanks is no different in this regard, so shield the last } of each \thanks
% that ends a line with a % and do not let a space in before the next \thanks.
% Spaces after \IEEEmembership other than the last one are OK (and needed) as
% you are supposed to have spaces between the names. For what it is worth,
% this is a minor point as most people would not even notice if the said evil
% space somehow managed to creep in.

% The paper headers
\markboth{Journal of \LaTeX\ Class Files,~Vol.~14, No.~8, August~2015}%
{Shell \MakeLowercase{\textit{et al.}}: Bare Demo of IEEEtran.cls for IEEE Journals}
% The only time the second header will appear is for the odd numbered pages
% after the title page when using the twoside option.
% 
% *** Note that you probably will NOT want to include the author's ***
% *** name in the headers of peer review papers.                   ***
% You can use \ifCLASSOPTIONpeerreview for conditional compilation here if
% you desire.

% If you want to put a publisher's ID mark on the page you can do it like
% this:
%\IEEEpubid{0000--0000/00\$00.00~\copyright~2015 IEEE}
% Remember, if you use this you must call \IEEEpubidadjcol in the second
% column for its text to clear the IEEEpubid mark.

% use for special paper notices
%\IEEEspecialpapernotice{(Invited Paper)}

% make the title area
\maketitle

% As a general rule, do not put math, special symbols or citations
% in the abstract or keywords.
\begin{abstract}
Automated waterway environment perception is crucial for enabling unmanned surface vessels (USVs) to understand their surroundings and make informed decisions. Most existing waterway perception models primarily focus on instance-level object perception paradigms (e.g., detection, segmentation). However, due to the complexity of waterway environments, current perception datasets and models fail to achieve global semantic understanding of waterways, limiting large-scale monitoring and structured log generation. With the advancement of vision-language models (VLMs), we leverage image captioning to introduce WaterCaption, the first captioning dataset specifically designed for waterway environments. WaterCaption focuses on fine-grained, multi-region long-text descriptions, providing a new research direction for visual geo-understanding and spatial scene cognition. Exactly, it includes 20.2k image-text pair data with 1.8 million vocabulary size. Additionally, we propose Da Yu, an edge-deployable multi-modal large language model for USVs, where we propose a novel vision-to-language projector called Nano Transformer Adaptor (NTA). NTA effectively balances computational efficiency with the capacity for both global and fine-grained local modeling of visual features, thereby significantly enhancing the model’s ability to generate long-form textual outputs. Da Yu achieves an optimal balance between performance and efficiency, surpassing state-of-the-art models on WaterCaption and several other captioning benchmarks. The project is available at https://github.com/GuanRunwei/WaterCaption.
\end{abstract}

% Note that keywords are not normally used for peerreview papers.

% For peer review papers, you can put extra information on the cover
% page as needed:
% \ifCLASSOPTIONpeerreview
% \begin{center} \bfseries EDICS Category: 3-BBND \end{center}
% \fi
%
% For peerreview papers, this IEEEtran command inserts a page break and
% creates the second title. It will be ignored for other modes.
\IEEEpeerreviewmaketitle

\section{Introduction}

\IEEEPARstart{W}{ith} the rapid advancement of deep learning, waterway perception based on deep neural networks has achieved remarkable progress. Currently, the development of unmanned surface vessels (USVs)-based waterway perception primarily focuses on tasks such as object detection \cite{ribeiro2017data}, semantic segmentation \cite{guan2024asy}, object tracking \cite{yao2025track}, simultaneous localization and mapping (SLAM) \cite{cheng2021we}, and scene understanding \cite{guan2023achelous, guan2023achelous++, guan2025watervg}. However, due to the inherent complexity of inland waterways, the difficulty of monitoring has significantly increased.

\begin{figure}
    \includegraphics[width=0.99\linewidth]{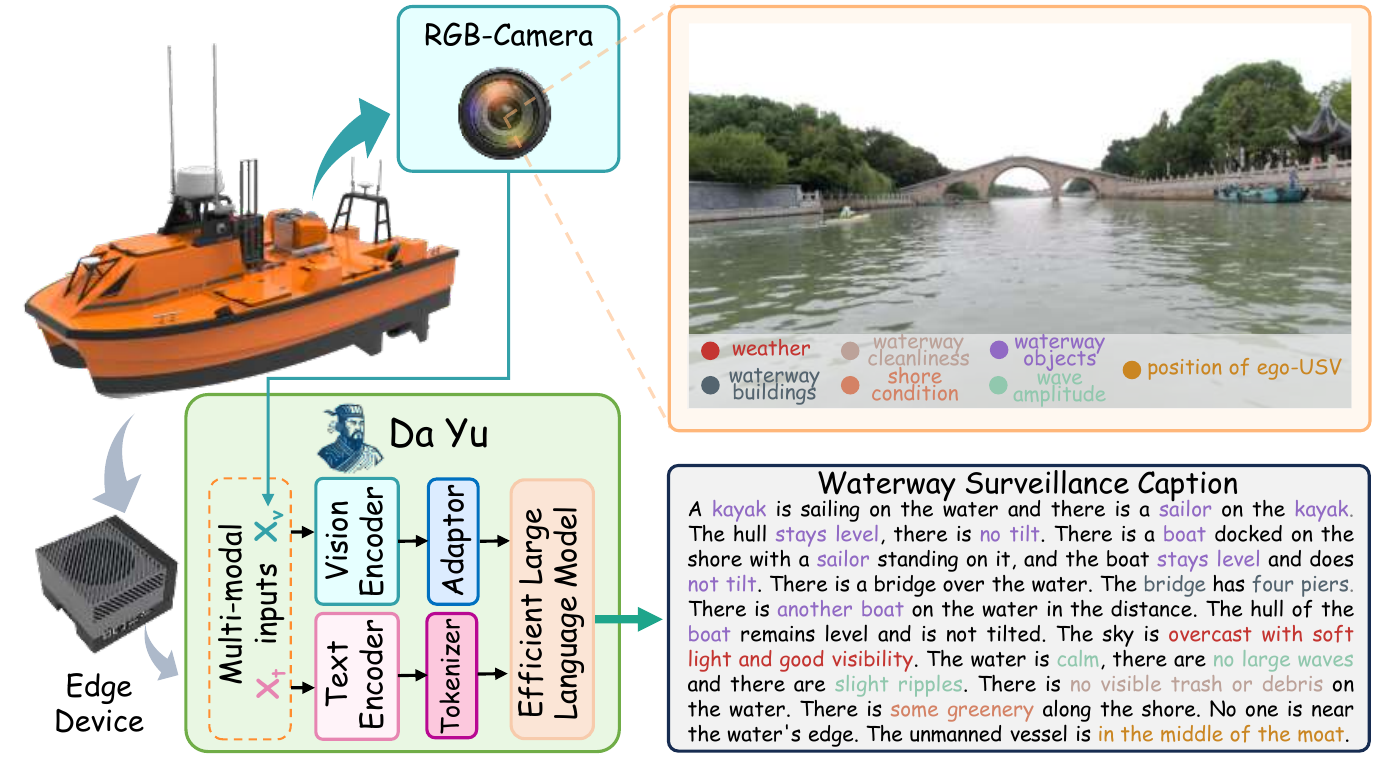}
    \vspace{-3mm}
    \caption{Overview of the USV-oriented waterway automatic captioning.}
    \label{fig:overview}
\end{figure}

To ensure safe and intelligent navigation, USVs are required to perceive and interpret multiple elements in real time, including navigational objects, water surface conditions, weather, shoreline structures, and pollutants, while also generating real-time regulatory logs to support intelligent path planning and obstacle avoidance. Nevertheless, existing perception systems remain largely constrained to instance-level paradigms, such as detection and segmentation, and have yet to explore global, semantic-level understanding of the waterway environment.

\begin{table*}
\setlength\tabcolsep{15.0pt}
\caption{Current visual grounding datasets and their characteristics.}
\vspace{-3mm}
\centering
\label{tab:vg_datasets}
\begin{tabular}{l|ccccccccccc}  
\toprule   
    \multirow{2}[2]{*}{\textbf{Datasets}} & \multirow{2}[2]{*}{\textbf{Source (Year)}} & \multirow{2}[2]{*}{\textbf{Domain}} & \textbf{Nb. Caps} & \multirow{2}[2]{*}{\textbf{Vocab Size}} & \textbf{Nb. Words} & \multirow{2}[2]{*}{\textbf{Nb. Images}} \\   
    & & & \textbf{(per image)} &  & \textbf{(per Cap)} \\
    \hline
    COCO \cite{chen2015microsoft} & Arxiv 2015 & Generic & 5 & 27K & 10.5 & 132K \\
    Flickr30K \cite{plummer2015flickr30k} & ICCV 2015 & Generic & 5 & 18K & 12.4 & 31K \\
    CC12M \cite{changpinyo2021conceptual} & CVPR 2021 & Generic & 1 & 523K & 20.0 & 12.4M \\
    SBU Captions \cite{ordonez2011im2text} & NeurlPS 2011 & Generic & 1 & 238K & 12.1 & 1M \\
    \hline
    VizWiz \cite{gurari2018vizwiz} & CVPR 2018 & Assistive & 5 & 20K & 13.0 & 70K \\
    CUB-200 \cite{reed2016learning} & CVPR 2016 & Birds & 10 & 6K & 15.2 & 12K \\
    Fashion Cap \cite{yang2020fashion} & ECCV 2020 & Fashion & 1 & 17K & 21.0 & 130K \\
    BreakingNews \cite{ramisa2017breakingnews} & TPAMI 2017 & News & 1 & 85K & 28.1 & 115K \\
    TextCaps \cite{sidorov2020textcaps} & ECCV 2020 & OCR & 5/6 & 44K & 12.4 & 28K \\

  \hline
  \textbf{WaterCaption (ours)} & 2025 & Waterway & 1 & 1.8M & 88.6 & 20.2K \\
\bottomrule  
\end{tabular}
\\
\smallskip
\footnotesize
\end{table*}

With the emergence of vision-language models (VLMs), perception paradigms have become increasingly flexible. Language can now serve as a supervisory signal to guide vision in tasks like visual grounding, while visual inputs can also be used to generate natural language descriptions through image captioning \cite{ghandi2023deep}. This opens up new possibilities for achieving comprehensive, interpretable, and semantically rich perception in complex waterway environments. 

Building upon this motivation, we propose WaterCaption, the first exploratory visual captioning dataset tailored for waterway perception. Unlike conventional datasets that focus solely on salient objects, WaterCaption encompasses a wide range of scene components, including navigational targets, water surface conditions, weather, pollution, and riverbanks. Furthermore, the annotations in existing general-purpose datasets (such as COCO \cite{chen2015microsoft} and Flickr30k \cite{plummer2015flickr30k}) are typically limited to object names and simple actions (such as “The boat is sailing on the water"), lacking descriptions of domain-specific attributes. As a result, the dataset is designed to support fine-grained, multi-region long-text descriptions, with strict requirements on textual coherence and logical consistency for long-form caption generation. WaterCaption is developed based on the large-scale autonomous surface driving dataset WaterScenes \cite{yao2024waterscenes}. It consists of 20,193 samples, covering four types of waterway scenarios, four weather conditions, and three illumination settings. Each caption includes detailed descriptions of 10 categories of navigational objects as well as various dynamic and static properties of the waterway environment, providing a comprehensive benchmark for studying long-text generation in complex multi-modal maritime scenes.

Correspondingly, considering the existence of communication blind spots in waterway environments, traditional cloud-based inference, with an average latency of 20 seconds over 4G or 5G networks, fails to meet the 8-second response time required by the Inland Collision Avoidance Regulations \cite{guard2018navigation} for the normal USVs. To address this limitation and enable real-time generation of waterway monitoring logs, we propose Da Yu$^1$\footnotetext[1]{$^1$Da Yu, is a legendary figure in ancient Chinese history renowned for his extraordinary efforts in controlling floods and protecting waterways. }, an edge-deployable multi-modal large language model (MLLM) for maritime monitoring. Da Yu fundamentally redefines onboard intelligence by synergistically integrating our novel Nano Transformer Adaptor (NTA), which eliminates the traditional trade-off between computational efficiency and environmental comprehension. Through its dual innovation of Multi-Head Pooling Attention and Dilated Grouped Convolution, the NTA enables linear-complexity processing of high-resolution visual streams while preserving pixel-level fidelity for critical maritime features (e.g., floating debris, vessel identification, and shoreline anomalies). Specifically, the NTA preserves the capability for global context modeling while reducing the parameter count by approximately 24\% compared to the convolution-based LDPv2 module \cite{chu2024mobilevlm2}. This breakthrough allows Da Yu to perform autonomous, cloud-free situational awareness of dynamic waterways, analysing complex waterway scenarios in real-time on embedded Jetson Orin platforms.

Based on the aforementioned content, the contributions of the paper are as follows:

\begin{enumerate}
    \item We construct the WaterCaption dataset, the first fine-grained, long-text image caption dataset specifically designed for waterway and maritime navigation scenarios. 
    \item We propose Da Yu, the first efficient MLLM tailored for the waterway navigation domain. Da Yu is capable of generating real-time waterway monitoring reports and achieves state-of-the-art performance.
    \item We design a novel image-to-text adaptor called Nano Transformer Adaptor. It possesses a global receptive field while effectively capturing local features within neighborhoods, achieving a well-balanced trade-off between accuracy and inference speed.
    \item We comprehensively analyse the performances of models with different paradigms and construct a full-scale benchmark for image captioning with long text.
\end{enumerate}

We organize the remaining content as follows. Section \ref{sec:related} introduces the related works; Section \ref{sec:dataset} presents our constructed WaterCaption dataset; Section \ref{sec:method} illustrates the proposed MLLM termed Da Yu; Section \ref{sec:experiments} shows the experiment results; Section \ref{sec:limitations} states the limitations of the research and future works; Section \ref{sec:conclusion} presents the conclusion.

% To address this limitation and enable real-time generation of waterway monitoring logs, we propose Da Yu, the first edge-deployable multi-modal large language model for maritime monitoring. Da Yu integrates complex environmental understanding capabilities directly into onboard edge devices of USVs, enabling autonomous situational awareness of dynamic waterway conditions without reliance on cloud computing.

\begin{figure*}
    \includegraphics[width=0.99\linewidth]{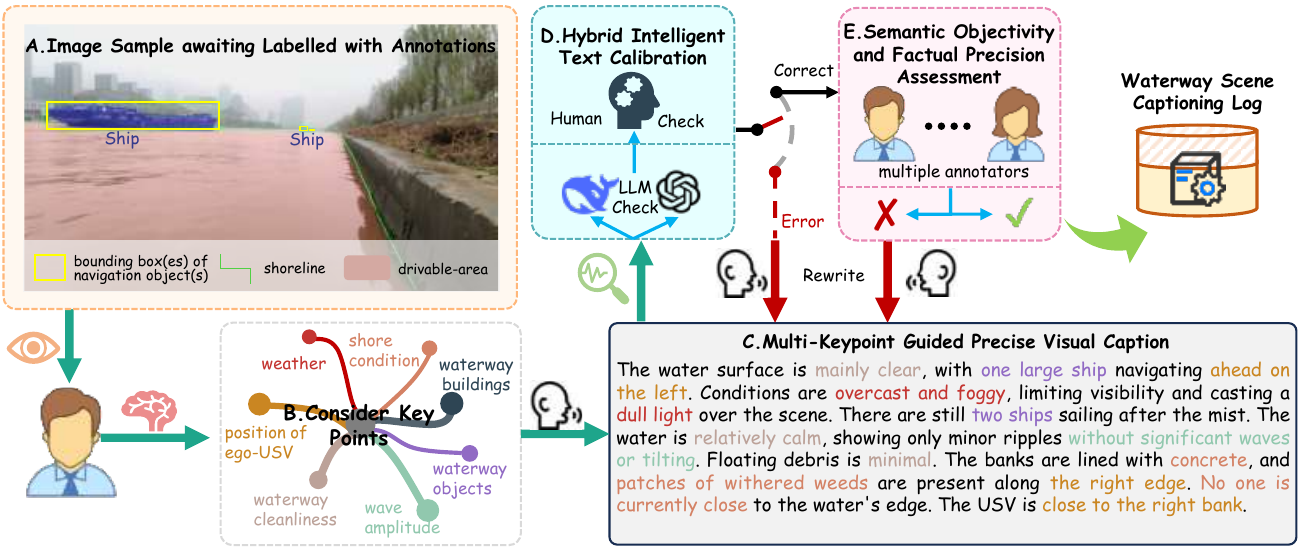}
    \vspace{-3mm}
    \caption{The process of captioning annotation for the WaterCaption dataset, which mainly contains four steps (A, B, C, D and E).}
    \label{fig:annotation_process}
\end{figure*}

\section{Related Works}
\label{sec:related}

\subsection{Image Captioning based on Vision-Language Models}
Recent advancements in VLMs have significantly improved the performance of image captioning tasks. Models such as CLIP \cite{radford2021learning}, ALBEF \cite{li2021align}, BLIP \cite{li2022blip}, and BLIP-2 \cite{li2023blip} have demonstrated remarkable capabilities in aligning visual and textual modalities. CLIP leverages contrastive learning to align image and text embeddings, enabling zero-shot and few-shot learning capabilities across various vision-language tasks. ALBEF introduces an "align before fuse" mechanism to address challenges in image-text alignment and interaction within the multi-modal domain, achieving notable performance improvements in tasks such as image-text retrieval and cross-modal generation. BLIP and BLIP-2 further enhance this approach by introducing a bootstrapping method to generate synthetic captions and filter noisy ones, achieving state-of-the-art results in tasks like image captioning and visual question answering. The integration of large language models (LLMs) with VLMs has opened new avenues for generating richer and more context-aware captions. For instance, models like LLaVA \cite{liu2023visual} and Qwen-VL \cite{yang2024qwen2} combine the strengths of CLIP and LLMs to handle complex visual-linguistic tasks. However, generating long-text captions for multiple regions within an image remains a significant challenge. Traditional VLMs often struggle with computational efficiency and scalability when handling long-text generation, as they require processing extensive multi-modal interactions. Moreover, these models typically have limited reasoning capabilities compared to multi-modal large language models (MLLMs), making it difficult to maintain coherence and semantic consistency over extended sequences. Another limitation of traditional VLMs is their reliance on object detection models, which primarily focus on foreground objects and often fail to provide precise descriptions of background elements \cite{yan2021task,cao2022vision,xian2022adaptive}. As noted in recent studies, this limitation arises because detection-based captioning models are trained on datasets with sparse annotations for background regions \cite{jiang2022double,yu2023comprehensive}, leading to incomplete scene descriptions \cite{zhang2023spt,yu2019multimodal}.

To address these challenges, lightweight MLLMs, such as the proposed Da Yu, have emerged as promising solutions. By optimizing model architecture and leveraging efficient training strategies, Da Yu demonstrates superior performance in generating coherent and context-aware long-text captions for multiple regions while maintaining computational efficiency.

\begin{figure*}
    \includegraphics[width=0.99\linewidth]{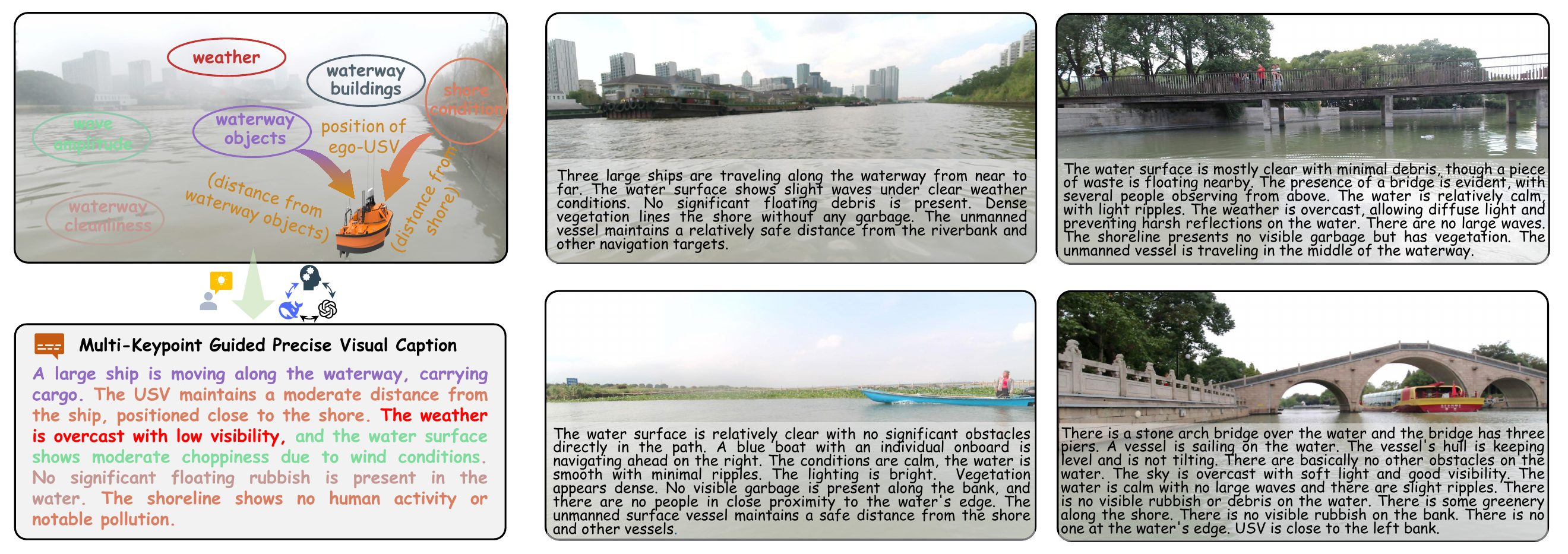}
    \vspace{-3mm}
    \caption{Samples in the WaterCaption dataset. The rightmost sample shows the caption text corresponding to different regions.}
    \label{fig:data_samples}
\end{figure*}

\subsection{Efficient Multi-Modal Large Language Model}
Recent research has witnessed a surge of interest in lightweight yet powerful MLLMs. These models, with parameters typically under 3B, are designed to efficiently handle vision-language tasks with limited computational resources. MobileVLM \cite{wu2024mobilevlm} and its enhanced version MobileVLM v2 \cite{chu2024mobilevlm2} focus on optimizing architectures and training methods to meet mobile application requirements. InternVL 2.5 (0.9B) \cite{chen2024internvl} leverages advanced training strategies like MPO on MMPR-v1.1, achieving impressive performance with relatively smaller visual and language encoders. Qwen-VL (0.9B) \cite{yang2024qwen2} combines a medium-sized visual encoder with a lightweight language model to effectively process visual inputs and generate textual outputs. MiniCPM-V \cite{yao2024minicpm}, built on CPM-Bee, inherits strong language generation capabilities and integrates visual understanding. TinyLLaVA \cite{zhou2024tinyllava} explores lightweight architectures and training optimizations to achieve high efficiency while maintaining performance. These models collectively push the boundaries of efficient multi-modal processing in resource-constrained environments.

However, these models, being pre-trained on general-purpose datasets, still exhibit limited understanding of maritime scenarios. To address this, and to enhance the perceptual capabilities of MLLMs in the maritime domain while fostering an intelligent perception ecosystem for waterborne transportation, we propose Da Yu, a lightweight MLLM designed for maritime perception and optimized for edge inference on USV platforms.

\subsection{Waterway Perception based on Deep Learning}
As a critical component of intelligent urban transportation systems, waterway perception is pivotal to maritime navigation and aquatic environmental conservation. Recent advances in deep learning have led to the widespread adoption of data-driven perception models for waterway monitoring. Prior works primarily focus on specialized tasks: object detection for maritime environments \cite{guan2023achelous}, free-space and shoreline segmentation \cite{guan2024mask}, and end-to-end panoptic perception frameworks unifying detection and segmentation \cite{guan2024asy}. These approaches predominantly rely on multi-sensor fusion for autonomous perception. While recent studies \cite{guan2025watervg,guan2024nanomvg} have introduced visual grounding models for intent-aware maritime object localization, the critical task of automated waterway surveillance and log generation, which is essential for navigational safety, remains underexplored due to the lack of dedicated datasets. To bridge this gap, we propose the first multi-regional Waterway Captioning Dataset (WaterCaption), capturing rich semantic descriptions of aquatic environments from USV-deployed RGB cameras in various waterway types. Building upon this, we introduce a novel MLLM called Da Yu specifically optimized for waterway captioning, demonstrating state-of-the-art capabilities in long-form text generation and complex scene understanding under challenging conditions.

\section{WaterCaption Dataset}
\label{sec:dataset}

Overall, the WaterCaption dataset comprises a total of 20,247 samples, including 14,135 samples in the training set, 2,019 in the validation set, and 4,039 in the test set. Each sample consists of two components: (1) an RGB image captured from a first-person perspective onboard an USV navigating through diverse waterway environments under varying weather conditions, lighting scenarios, and times of day; and (2) a fine-grained caption text describing the waterway scene, generated in the style of an automated navigation log based on the corresponding RGB image. Fig. \ref{fig:data_samples} shows the data samples in the WaterCaption dataset, where the rightmost panel presents the caption descriptions associated with distinct regions in the waterway image.

\subsection{Dataset Construction}
The WaterCaption dataset is built upon the WaterScenes dataset, a large-scale and high-quality benchmark for waterway perception. WaterScenes encompasses a wide variety of navigation scenarios, including diverse channel types, weather conditions, visibility levels, illumination settings, and time-of-day variations. Each image in the WaterScenes dataset is provided at a resolution of 1920 × 1080.

Fig. \ref{fig:annotation_process} presents the annotation process of WaterCaption dataset, which primarily contains five steps.

\textit{1) Image sample awaiting labelled with annotations:} Each image sample awaiting labelled is accompanied by rich annotations provided by WaterScenes dataset, including 2D bounding boxes of navigational objects, drivable area segmentation, and shoreline delineation. These annotations serve as reliable references to support annotators during the generation of fine-grained captions.

\textit{2) Consider Key Points:} In constructing waterway-oriented captions, we systematically consider seven key semantic regions (or focal elements) that comprehensively reflect the functional (navigability), safety-related (risk factors), and ecological (environmental health) attributes of the scene. These seven elements include: \textbf{(1) waterway objects}, which directly impact navigational safety and route planning in the channel; \textbf{(2) waterway buildings}, which reveal human interventions and functional demands on the waterway infrastructure; \textbf{(3) weather conditions}, which critically affect image visibility and perception; \textbf{(4) wave amplitude}, which influences the stability of vessel navigation; \textbf{(5) shore condition}, indicating ecological well-being, environmental pollution levels, and the intensity of anthropogenic activities; \textbf{(6) waterway cleanliness}, directly associated with environmental protection and maintenance; and \textbf{(7) the position of the ego-USV}, which ensures the USV remains in a relatively safe spatial relationship with navigation targets and shoreline boundaries. However, not all scenes contain all of the aforementioned seven key elements. In certain waterway scenarios, such as on open lakes or at sea where the shoreline is not visible, the shore condition element is not required to be included in the caption.

In summary, compared to general-purpose datasets such as COCO and Flickr30k, which primarily focus on object names and simple action states, the proposed WaterCaption dataset incorporates domain-specific attributes, including quantitative indicators, scene dynamics, as well as ecological and socio-environmental descriptors. These enriched descriptions address a critical gap in the current image captioning literature by introducing fine-grained, highly interpretable semantic information tailored for waterway scenarios.

\textit{3) Multi-Keypoint Guided Precise Visual Caption:} The annotation process of the WaterCaption dataset involved seven expert annotators with domain expertise in maritime transportation. In constructing the image captions, we first ensur that all seven key elements identified in Step 1 were adequately represented. Importantly, the annotation protocol is designed with consideration for how humans naturally describe complex scenes, employing diverse syntactic structures and varying the order of information presentation. For instance, salient objects may be described first, followed by contextual or relational details, often linked through subordinate clauses.

To preserve the naturalness and expressiveness of the language, we deliberately avoid imposing a fixed ordering scheme (e.g., left-to-right or near-to-far). Rigid spatial sequencing can reduce linguistic variability and may inadvertently encourage models to rely on positional heuristics rather than true semantic understanding. By incorporating syntactic diversity and randomized description order, we compel the model to focus on the semantic coherence and contextual relevance of the content. This design choice enhances the model’s compositional generalization and robustness in downstream reasoning and captioning tasks.

\textit{4) Hybrid Intelligent Text Calibration: }Based on the initially constructed captions, we employ a collaborative quality control pipeline involving two state-of-the-art large language models, including ChatGPT-4o \cite{achiam2023gpt} and DeepSeek-R1 \cite{liu2024deepseek}, working in tandem with human annotators for iterative grammar and semantic validation. This hybrid framework significantly enhances the linguistic consistency and domain correctness of the dataset by forming a closed-loop verification system that integrates multi-modal error detection with expert human judgment. Specifically, the dual-model approach enables complementary identification of general grammatical issues and domain-specific terminology inconsistencies, while human reviewers resolve contextual ambiguities and ensure semantic coherence. This automated-first, human-in-the-loop strategy strikes a balance between annotation efficiency and precision, yielding a high-quality, low-noise multi-modal alignment benchmark suitable for rigorous academic research and downstream applications.

\textit{5) Semantic Objectivity and Factual Precision Assessment: }To mitigate potential ambiguities and subjective inconsistencies in waterway caption, we adopt a consensus-based annotation strategy involving seven domain-expert annotators. Through collective deliberation, this approach systematically resolves semantic vagueness or interpretative bias that may arise from individual cognitive differences. By leveraging collective intelligence, the process ensures textual objectivity and factual accuracy, thereby providing ultra-low-noise training data for vision-language models and minimizing annotation-induced data bias to the greatest extent possible.

\begin{figure}[h]
    \includegraphics[width=0.99\linewidth]{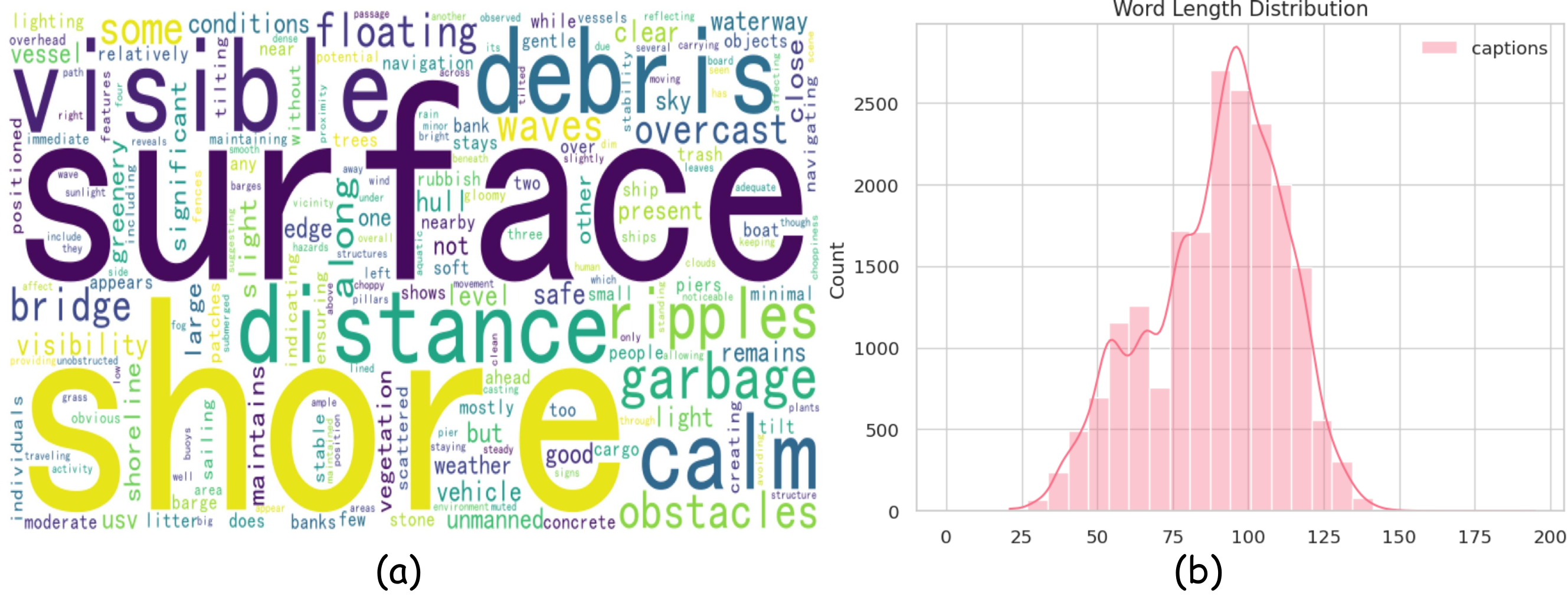}
    \vspace{-3mm}
    \caption{Statistics of captioning corpus in WaterCaption dataset. (a) presents the word cloud while (b) shows the distribution of caption length.}
    \label{fig:word_stat}
\end{figure}

\begin{figure}
    \includegraphics[width=0.999\linewidth]{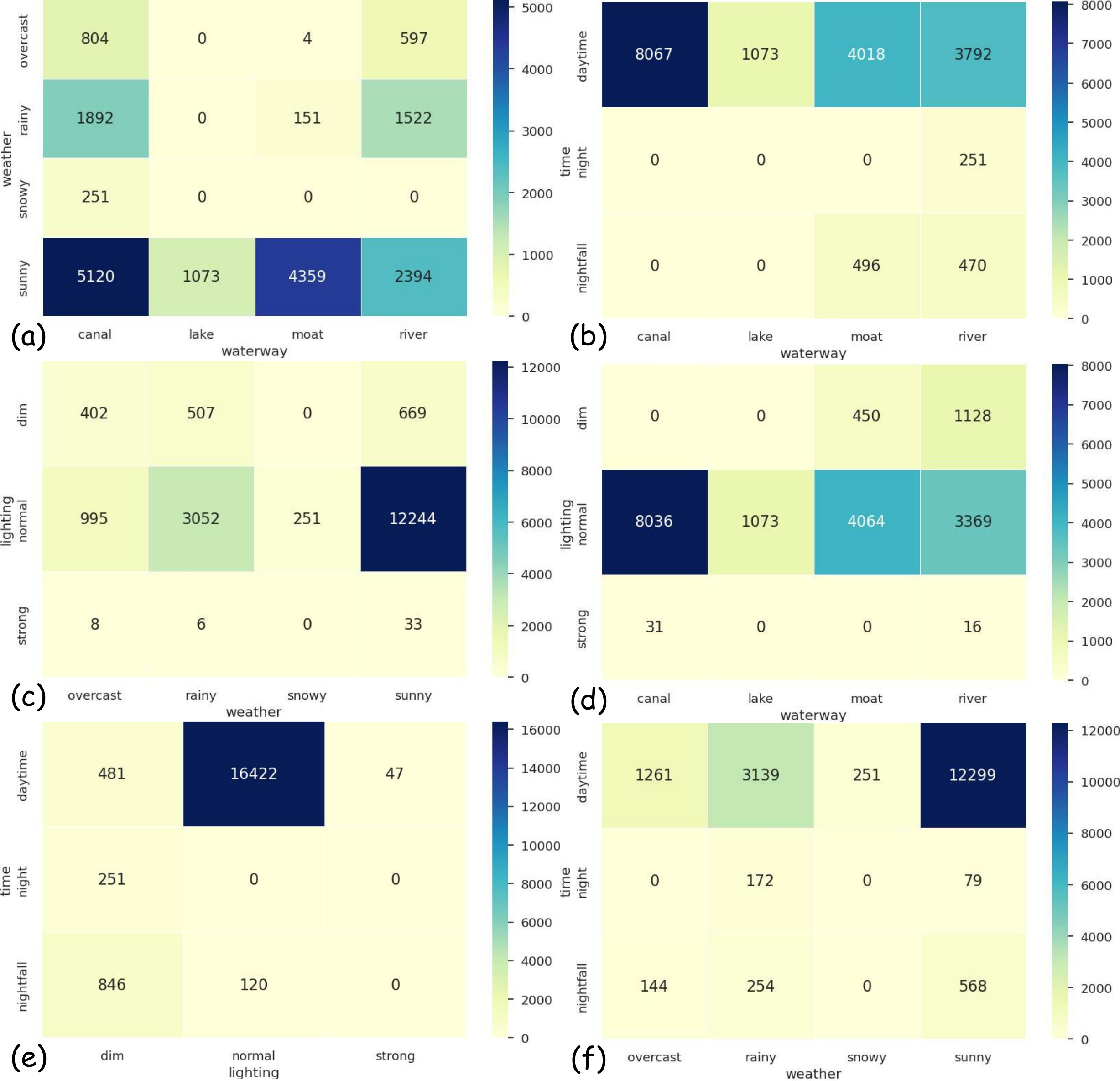}
    \vspace{-5mm}
    \caption{Pairwise statistics of environmental variables (time, waterway, weather and lighting) in WaterCaption dataset.}
    \label{fig:dataset_characteristics_map}
\end{figure}

\subsection{Dataset Statistics}

Fig. \ref{fig:word_stat} (a) illustrates the word cloud of the WaterCaption dataset after removing all stop words, articles, and other terms irrelevant to the environmental context. Fig. \ref{fig:word_stat} (b) presents the distribution of caption lengths, showing that most captions range between 75 and 100 words. Compared to existing captioning datasets, WaterCaption features significantly longer captions, thereby posing greater challenges for text generation.

\begin{figure}
\centering
    \includegraphics[width=0.85\linewidth]{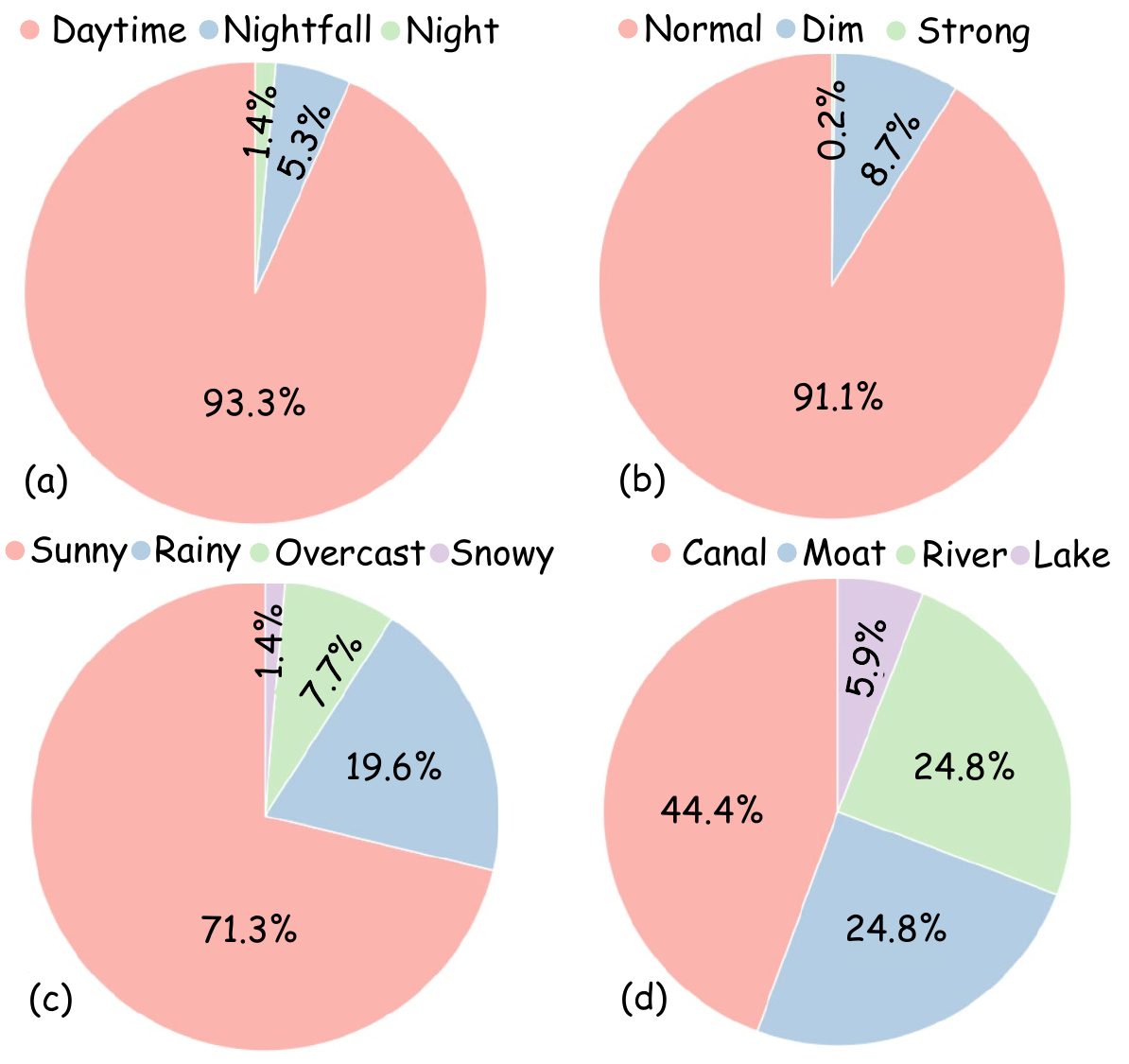}
    \vspace{-3mm}
    \caption{The proportion of environmental variables for WaterCaption, including (a) time of day, (b) lighting, (c) weather, (d) waterway.}
    \label{fig:pie_chat}
\end{figure}

Fig. \ref{fig:dataset_characteristics_map} illustrates the co-occurrence frequency of pairwise combinations of different environmental variables in waterway scenes. These variables include four major categories: waterway type, time of day (daytime), lighting conditions, and weather. The high frequency and diversity of these combinations highlight the rich environmental variability captured in WaterCaption, which in turn implies a high degree of caption diversity. Besides, Fig. \ref{fig:pie_chat} presents the proportation of environmental variables for WaterCaption dataset.

Moreover, to quantitatively evaluate the textual richness of the dataset, we propose two metrics: Tri-Grams Complexity (\textit{TGC}) and Complex Word Ratio (\textit{CWR}), which measure syntactic and semantic variability and lexical sophistication, respectively. Specifically, first, the ratio of unique tri-grams in the corpus quantifies the variability of word sequences at contextual scales with three words. Higher values indicate richer syntactic and semantic patterns. Eq. \ref{eq:tri_gram} shows the calculation method of \textit{TGC}.

\begin{equation}
    TGC = \frac{U_3}{T_3},
    \label{eq:tri_gram}
\end{equation}
where $U_3$ denotes the number of unique tri-grams and $T_3$ is the total number of tri-grams.

Secondly, we define the Complex Word Ratio (\textit{CWR}) as the proportion of words exceeding seven characters in length within the corpus. This metric serves as an indicator of lexical sophistication and the density of domain-specific terminology. Eq. \ref{eq:cwr} presents the calculation of CWR.

\begin{equation}
    CWR = \frac{N_{complex}}{N_{total}},
    \label{eq:cwr}
\end{equation}
where $N_{complex}$ denotes the number of words with length more than 7 while $N_{total}$ is the total number of words in the corpus.

\begin{table}[h]
\setlength\tabcolsep{2.0pt}
\caption{Caption complexity of image captioning datasets.}
\vspace{-3mm}
\centering
\label{tab:datasets_char_compare}
\begin{tabular}{l|cc}  
\toprule   
    \textbf{Datasets} & \textbf{Tri-Grams Complexity $\uparrow$} & \textbf{Complex Word Ratio $\uparrow$} \\
    \midrule
    COCO \cite{chen2015microsoft} & 0.528 & 0.240 \\
    Flickr30K \cite{plummer2015flickr30k} & \textbf{0.694} & 0.257 \\

  \midrule
  \textbf{WaterCaption (ours)} & 0.535 & \textbf{0.388}  \\
\bottomrule  
\end{tabular}
\\
\smallskip
\footnotesize
\end{table}

Table \ref{tab:datasets_char_compare} presents a quantitative comparison of different image captioning datasets in terms of TGC and CWR metrics. Our proposed WaterCaption dataset demonstrates a TGC value comparable to COCO but slightly lower than Flickr30K. However, in terms of CWR, the domain-specific, fine-grained, and multi-region WaterCaption dataset exhibits a significantly higher proportion of complex vocabulary than the general-domain datasets COCO and Flickr30K, highlighting the unique research value and linguistic richness of WaterCaption.

\begin{figure*}
    \includegraphics[width=0.99\linewidth]{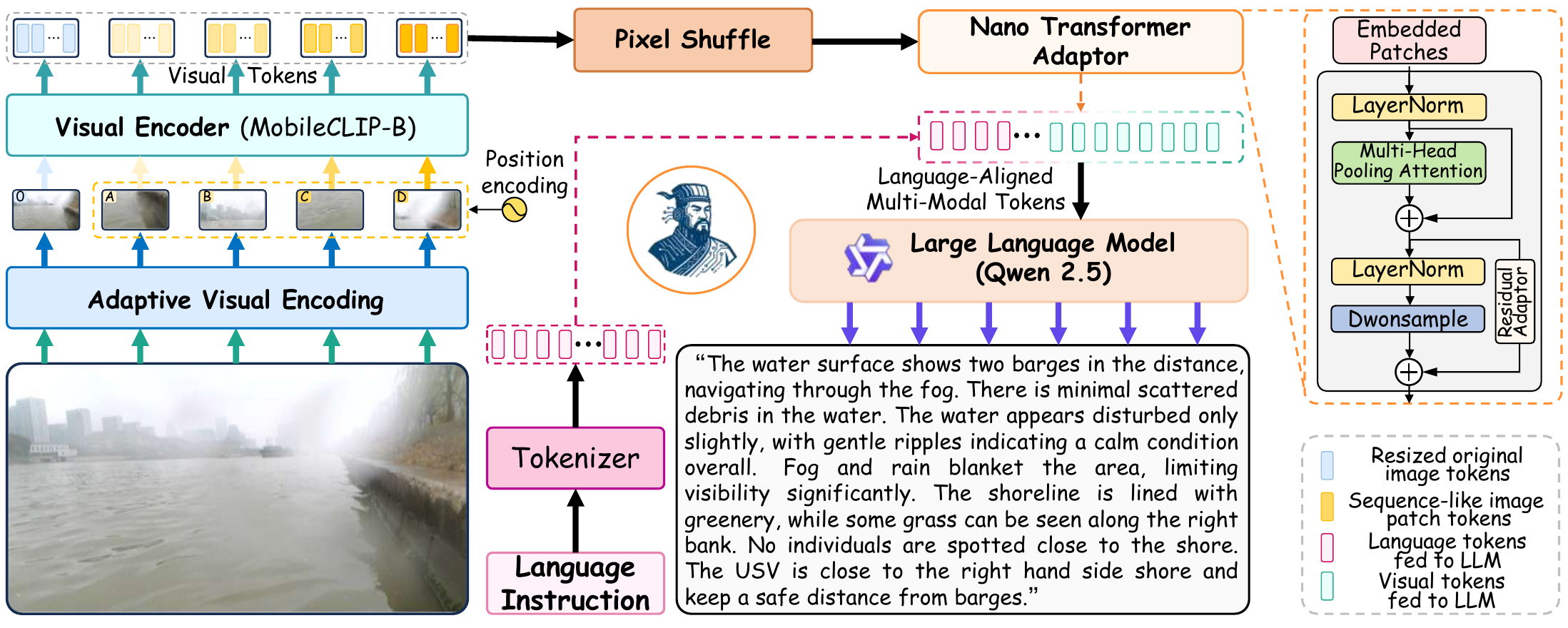}
    \vspace{-3mm}
    \caption{The architecture of our proposed multi-modal large language model: Da Yu\includegraphics[height=1.5em]{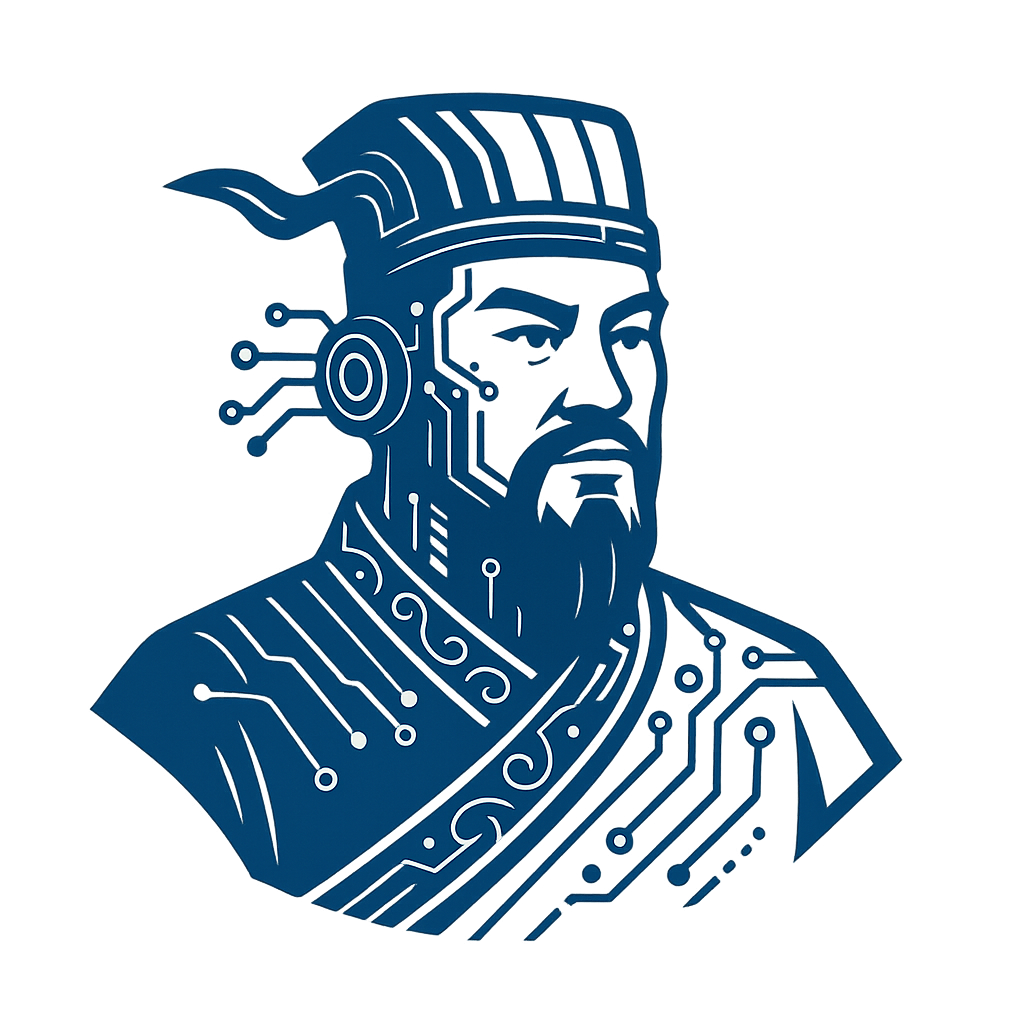}. Da Yu model mainly includes a visual encoder followed by the pixel shuffle operation, our proposed Nano Transformer Adaptor (NTP) and a large language model (Qwen 2.5)}
    \label{fig:model}
\end{figure*}

\section{Methodology}
\label{sec:method}

In this section, we introduce our proposed MLLM, Da Yu, as illustrated in Fig. \ref{fig:model}. Consistent with most MLLM architectures, Da Yu accepts two inputs—an RGB image and a natural language instruction, and generates a language-based output. The overall framework of Da Yu comprises three primary components: a visual encoder, a Nano Transformer Adaptor, and a pretrained LLM. Specifically, the RGB image is first processed through an adaptive visual encoding module to generate a sequence of visual patches. These patches are then forwarded to a pretrained visual encoder to produce visual tokens. Subsequently, the tokens are spatially rearranged using a pixel-shuffle operation and projected into the LLM-compatible embedding space via our proposed Nano Transformer Adaptor. The resulting visual embeddings are concatenated with the tokenized language instruction, forming a unified input sequence. This fused multi-modal input is then passed to the pretrained LLM to generate the final language output. The details of each module are elaborated in the subsequent sections.

\subsection{Adaptive Visual Encoding}
We adopt an adaptive image encoding method inspired by \cite{yao2024minicpm}. First, the input image is adaptively sliced into multiple patches. If the image pixel exceeds a threshold (e.g., 448×448), it is split into $N$ slices. The slices are arranged in a grid pattern with $m$ rows and $n$ columns ($m×n=N$), where the ($m$, $n$) combination is chosen to minimize the aspect ratio difference with the original image.
When the image pixel is within the threshold, it is resized to the target size while maintaining its aspect ratio. For slice size adjustment, each sliced image may differ in size from the Visual Transformer (ViT) pre-training region size ($W_v×H_v$) in MobileCLIP-B \cite{vasu2024mobileclip}. So, we proportionally adjust the size of each slice as $adj\_slice$:

\begin{equation}
    adj\_slice = ori\_slice \times \frac{W_v}{ori\_width} \times \frac{H_v}{ori\_height},
\end{equation}
where $ori\_slice$ denotes the original image slice before any adjustment. $W_v$ and $H_v$ denote the width and height of the ViT pre-training region. $ori\_width$ and $ori\_height$ denote the width and height of the original image slice before adjustment.

\subsection{Visual Encoder}
We adopt MobileCLIP \cite{vasu2024mobileclip} as the visual encoder, a lightweight and efficient model distilled from the image captioning framework CoCa \cite{yucoca} and OpenAI-CLIP \cite{radford2021learning}. MobileCLIP inherits CLIP’s capability to capture global, low-frequency, coarse-grained visual label semantics, while simultaneously leveraging the fine-grained, low-frequency visual details learned from CoCa. Compared to SigLIP \cite{zhai2023sigmoid} and the OpenAI-CLIP model, MobileCLIP not only has a smaller parameter footprint but also demonstrates superior image representation performance. 

Subsequently, we employ an effective compression strategy, pixel shuffle (space-to-depth) to rearrange spatial features into additional channel dimensions. This operation reduces the spatial resolution while increasing the representational density. Specifically, it decreases the total number of visual tokens by a factor of $r^2$, where $r$ denotes the shuffle rate. However, using a larger shuffle rate compresses a broader spatial area into a single token, which may adversely affect tasks requiring precise localization. To balance compression efficiency and spatial fidelity, we set $r = 2$. This configuration significantly reduces token count, thereby lowering the computational cost of attention and enhancing long-context modeling capabilities.

\subsection{Nano Transformer Adaptor}
To project visual tokens into the language feature space, we propose the Nano Transformer Adaptor (NTA), a lightweight adaptor module with linear complexity that incorporates attention-based dynamic encoding. NTA effectively captures key regional features while reducing redundant computation. Compared to convolution-only adaptors, NTA achieves superior modeling of long-range cross-region dependencies through its global attention mechanism, all while maintaining comparable computational efficiency. Concurrently, NTA enhances the model's ability to reconstruct high-frequency information of the image and improves the long caption generation. This design provides a balanced solution in terms of efficiency, accuracy, and interpretability for multi-granularity and multi-object image captioning tasks.

\begin{figure}[h]
    \includegraphics[width=0.99\linewidth]{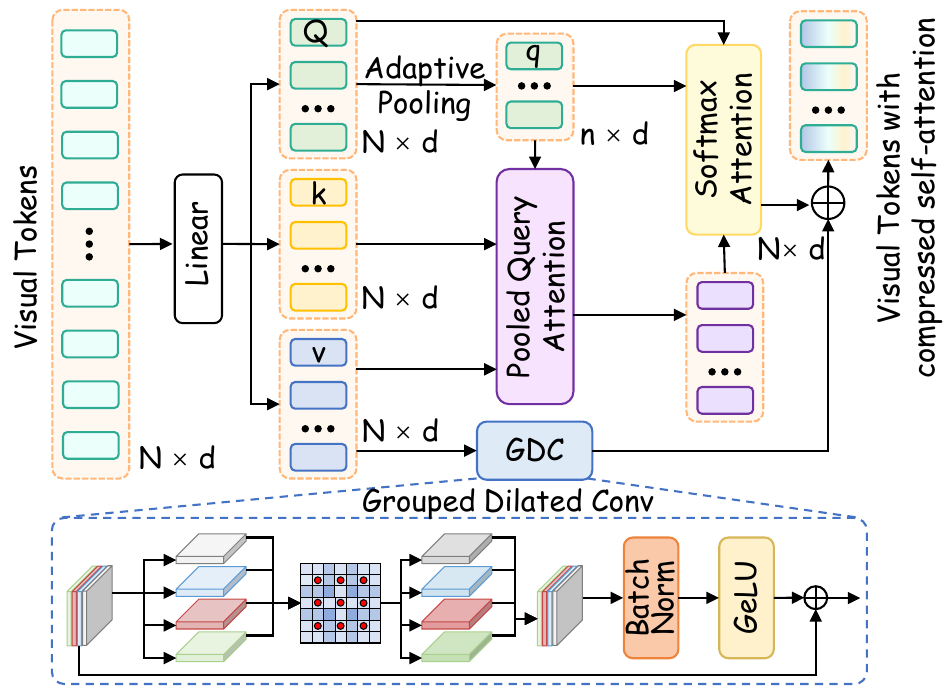}
    \vspace{-3mm}
    \caption{The structure of Pooling Attention in Nano Transformer Adaptor.}
    \label{fig:nta}
\end{figure}

The whole structure of NTA is presented in the upper right corner area of Fig. \ref{fig:model}. Given the sequence of input visual token $f \in \mathbb{R}^{N \times d}$, it first go through a layer-norm module:

\begin{equation}
    f_v = \mathtt{LayerNorm}(f), f_v \in \mathbb{R}^{N \times d}.
    \label{eq:layernorm}
\end{equation}

Here, we obtain the normalized visual feature $f_v$, which serves as the input to the multi-head pooling attention module.

As illustrated in Fig.~\ref{fig:nta}, the multi-head pooling attention mechanism first derives the query ($Q$), key ($K$), and value ($V$) feature vectors through a linear projection with a channel expansion ratio of three:

\begin{equation}
    Q, K, V = \mathtt{Linear}(f_v), \{Q, K, V\} \in \mathbb{R}^{N \times d}.
    \label{eq:qkv}
\end{equation}

% Inspired by \cite{han2024agent}, to reduce the redundant information from $Q$ while aggregating the information from $K$ and $V$, we first adopt adaptive pooling and get $q \in \mathbb{R}^{n \times d}$. Here, $n$ is a self-defined hyper-parameter, which is smaller than $N$. Then, in pooled query attention $\mathtt{PQA(\cdot)}$, $q$ is matrix multiplied with $K$ to calculate the similarity between them and obtain an attention score matrix $S$. The matrix $S$ is multiplied with $V$ and obtain the attention-based aggregated feature $A$. Subsequently, we adopt $q$ as key while $A$ as the value to broadcast the global information to the $Q$, in the softmax attention $\mathtt{Attn(\cdot)}$. Moreover, to enhance the reduced feature restoration in the pooling operation, we design a grouped dilated convolution $\mathtt{GDC}(\cdot)$ as the residual path to the final output $\hat{f}_v$ with extremely low parameter cost. The whole process is shown in Equation \ref{eq:pool_attn} and \ref{eq:pool_attn_residual}.
Inspired by the work in \cite{han2024agent}, and to mitigate redundancy in the query $Q$ while effectively aggregating contextual information from $K$ and $V$, we first apply an adaptive pooling operation to obtain a compressed query representation $q \in \mathbb{R}^{n \times d}$, where $n$ is a user-defined hyperparameter satisfying $n < N$. 

In the pooled query attention module $\mathtt{PQA(\cdot)}$, we compute the attention score matrix $S$ by performing a matrix multiplication between $q$ and $K$. The resulting matrix $S$ is then used to weight the value matrix $V$, yielding the attention-based aggregated feature $A$. Next, to propagate the global contextual information from the pooled space back to the original query space, we employ a softmax attention module $\mathtt{Attn(\cdot)}$, where $q$ serves as the key and $A$ as the value, and the attention is applied to the original $Q$. To further enhance the representation capability, especially to recover potential detail loss from the pooling operation, we introduce a lightweight residual path utilizing grouped dilated convolution $\mathtt{GDC}(\cdot)$, which contains the combination of a grouped dilated convolution, batch normalization and GeLU activation. Grouped dilated convolution efficiently enriches the final output $\hat{f}_v$ with minimal additional parameter cost. The complete procedure is formally described in Equations~\ref{eq:pool_attn} and~\ref{eq:pool_attn_residual}.

\begin{align}
& \left\{
    \begin{aligned}
    \label{eq:pool_attn}
        & q = \mathtt{AdaPool}(Q), q \in \mathbb{R}^{n \times d},  \\
        & S = q \cdot K^T, S \in \mathbb{R}^{n \times N}, \\
        & A = \mathtt{PQA}(S, V) = \frac{S}{\sqrt{d}}\cdot V, A \in \mathbb{R}^{n \times d}, \\
        & \hat{f}_v = \mathtt{Attn}(Q, q, A) = \frac{Q \cdot q^T}{\sqrt{d}}\cdot A + \mathtt{GDC}(V), \hat{f}_v \in \mathbb{R}^{N \times d},       
    \end{aligned}
\right.
\end{align}

\begin{equation}
    \tilde{f_v} = \hat{f}_v + f_v, \tilde{f_v} \in \mathbb{R}^{N \times d},
 \label{eq:pool_attn_residual}
\end{equation}
where $d$ represents the dimensionality of the attention components $S$, $Q$, and $q$. The proposed pooling attention module achieves a linear computational complexity of \( O(Nnd) \), which is substantially lower than that of conventional self-attention, whose complexity is \( O(N^2 d) \), while still maintaining the capacity for global context modeling. In parallel with the multi-head pooling attention mechanism, a residual connection is incorporated into the output $\hat{f}_v$, following the architectural design principles of the standard Transformer block.

Subsequently, to align the dimensionality of visual tokens with that of the language embedding space, a layer normalization followed by a downsampling convolution is applied. In parallel, a convolution-based residual adaptor is integrated into the final output to enhance representation fidelity. The overall process is formalized in the following equation:

\begin{align}
    & f_v^{LE} = \mathtt{LayerNorm}(\mathtt{Conv}_{\mathtt{d}}(\tilde{f_v})) + \mathtt{Conv}_{\mathtt{r}}(\tilde{f_v}),
\end{align}
where $\mathtt{Conv}_{\mathtt{d}}(\cdot)$ denotes the convolution-based downsampling operation while $\mathtt{Conv}_{\mathtt{r}}$ denotes the module of convolution-based residual adaptor.

\begin{figure}[!t]
    \includegraphics[width=0.99\linewidth]{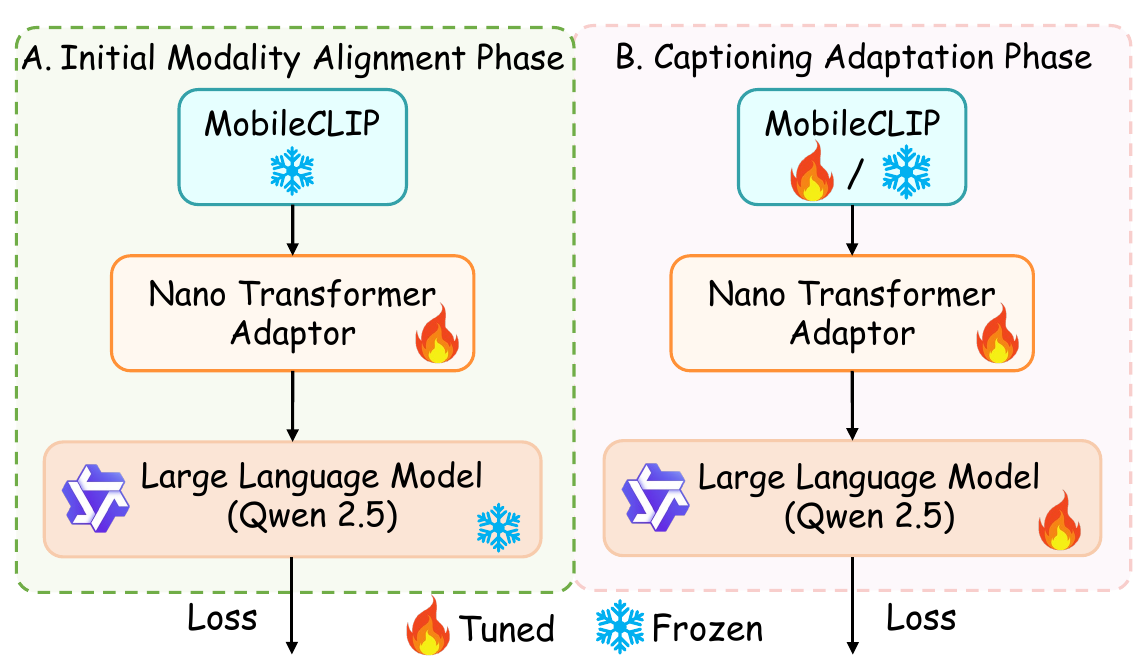}
    \vspace{-3mm}
    \caption{The overview of progressive two-stage training strategy.}
    \label{fig:two_stage_training}
\end{figure}

\subsection{Large Language Model}
We employ the Qwen 2.5 series \cite{yang2024qwen2}, specifically the 0.5B and 1.5B parameter variants as the LLM of Da Yu for waterway-oriented image captioning. Qwen 2.5 offers strong generalization with lightweight architecture, making it well-suited for real-time, edge-deployable scenarios.

Language-aligned multi-modal token sequence is processed by Qwen 2.5 in an autoregressive manner to generate fine-grained, domain-specific descriptions. Through lightweight instruction tuning on waterway-specific data, the model captures critical elements such as navigational entities, environmental attributes, and spatial semantics with high precision. This design ensures a balance between model expressiveness and computational efficiency, enabling high-quality caption generation under real-world deployment constraints.

\subsection{Training Procedure and Objective}
\textbf{Training Procedure:} We adopt a progressive two-stage training strategy as Fig. \ref{fig:two_stage_training} presents. First, we note that the visual encoder, MobileCLIP \cite{vasu2024mobileclip}, is built upon a pretrained model that has already captured rich visual representations. Second, to preserve the general linguistic knowledge and generation capabilities inherent in the pretrained large language model Qwen 2.5, we follow the training paradigm of LLaVA \cite{liu2023visual}. In the first stage of fine-tuning, we freeze the parameters of both the visual encoder and the large language model, and train only the Nano Transformer Adaptor. This facilitates effective alignment and fusion between the visual and language modalities, enabling the model to adapt to the image captioning task.

However, the parameter freezing in the first stage inevitably constrains the model’s representational capacity. Once the model acquires a preliminary understanding of the task, we proceed to the second stage by unfreezing the large language model for further training. This enhances the model’s ability to capture task-specific information. Moreover, since the MobileCLIP’s pretraining objective emphasizes low-frequency, high-level semantic features while being less sensitive to high-frequency details, we unfreeze the final block of MobileCLIP. This adjustment allows the Da Yu model to better capture and interpret the complex semantics and logical relationships required for fine-grained image captioning, ultimately improving overall performance.

\textbf{Training Objective:} Similar to MiniCPM-V \cite{yao2024minicpm} and MobileVLM \cite{wu2024mobilevlm}, we adopt the Causal Language Modeling (CLM) loss as the training objective for Da Yu, which is presented in Equation \ref{eq:clm}. CLM loss is specifically designed for training autoregressive language models and is widely used in generation tasks. Its core principle is to predict the probability distribution of the next token based on the preceding context, and to optimize the model parameters by maximizing the log-likelihood of the ground-truth next token.

\begin{equation}
    L_{CLM} = -\frac{1}{T} \sum_{t=1}^{T} \log p_{\theta}(x_t | x_{1:t-1}),
    \label{eq:clm}
\end{equation}
where $T$ represents the length of textual sequence. $x_t$ denotes the $t$-th character in the sequence. $x_{1:t-1}$ denotes the subsequence from the beginning of the sequence to the $t-1$th word or character, which serves as the context. $p_{\theta}(x_t | x_{1:t-1})$ is the probability that the model predicts that the $t$-th word is a given the context, and $\theta$ represents the parameters of the model.

\begin{table*}
    \setlength\tabcolsep{3.7pt}
    \caption{Overall performances on WaterCaption dataset.}
    \vspace{-3mm}
    \label{tab:benchmark_compare}
    \centering
    \begin{tabular}{c|c|c|ccccccccc}
    \toprule
       \textbf{Models} & \textbf{LLM} & \textbf{Params} & \textbf{ROUGH-1} & \textbf{ROUGH-2} & \textbf{ROUGH-L} & \textbf{BLEU-1} & \textbf{BLEU-2} & \textbf{BLEU-3} & \textbf{METEOR} & \textbf{CIDEr} & \textbf{GPT-Score} \\
    \hline
      DLCT  & - & - & - & - & 0.011 & - & - & - & 0.004 & - & - \\
      COS-Net & - & - & - & - & 0.017 & - & - & - & 0.003 & - & - \\
      \hline
      Oscar-B & - & 0.2B & - & - & 0.035 & - & - & - & 0.013 & - & 0.13 \\
      Oscar-L & - & 0.6B & - & - & 0.050 & - & - & - & 0.018 & - & 0.15 \\
      BLIP & - & 0.3B & 0.108 & 0.014 & 0.074 & 0.0002 & 0.000 & - & 0.024 & 0.00 & 0.18 \\
      BLIP-2 & - & 1.2B & 0.119 & 0.053 & 0.089 & 0.001 & 0.000 & - & 0.033 & 0.00 & 0.18 \\
      \hline
      InternVL 2.5 & Vicuna & 0.9B & 0.52 & 0.23 & \textbf{0.35} & 0.48 & \textbf{0.33} & 0.23 & \textbf{0.40} & 0.10 & 0.66 \\
      \textbf{Da Yu-S (ours)} & Qwen 2.5(0.5B) & 0.7B & \textbf{0.53} & \textbf{0.25} & \textbf{0.35} & \textbf{0.49} & \textbf{0.33} & 0.24  & \textbf{0.40} & \textbf{0.11} & \textbf{0.69}\\
      \hline
      MobileVLM & MobileLLaMA & 1.4B & 0.50 & 0.22 & 0.31 & 0.43 & 0.31 & 0.22 & 0.38 & 0.09 & 0.70 \\
      MobileVLM v2 & MobileLLaMA & 1.4B & 0.53 & 0.24 & 0.34 & 0.49 & 0.34 & 0.24 & 0.41 & 0.11 & 0.74 \\
      InternVL 2.5 & Vicuna & 1.9B & 0.55 & \textbf{0.28} & \textbf{0.37} & \textbf{0.50} & 0.34 & 0.26 & 0.43 & 0.13 & \textbf{0.81} \\
      \textbf{Da Yu-B (ours)} & Qwen 2.5(1.5B) & 1.7B & \textbf{0.57} & 0.27 & 0.36 & \textbf{0.50} & \textbf{0.36} & \textbf{0.27} & \textbf{0.43} & \textbf{0.13} & 0.79\\
      \hline
      Qwen-VL 2.5 & Qwen 2.5 & 3B & 0.58 & \textbf{0.29} & \textbf{0.38} & 0.50 & 0.36 & 0.27 & 0.45 & \textbf{0.15} & 0.81 \\
      MiniCPM-V 2.0 & MiniCPM & 2.8B & 0.57 & \textbf{0.29} & 0.34 & 0.48 & \textbf{0.39} & 0.27 & 0.42 & 0.12 & 0.78\\
      \textbf{Da Yu-L (ours)} & Qwen 2.5(3B) & 3.2B & \textbf{0.60} & \textbf{0.29} & \textbf{0.38} & \textbf{0.53} & 0.37 & \textbf{0.28} & \textbf{0.46} & \textbf{0.15} & \textbf{0.83} \\
    \bottomrule
    \end{tabular}
\end{table*}

\section{Experiments}
\label{sec:experiments}

\subsection{Implementation Details}

\textbf{Dataset Settings:} To evaluate the performances of our proposed Da Yu model on the generation of image captioning, we include our proposed WaterCaption dataset as the benchmark, which includes 20.2K image-captioning pairs under four types of waterway channels. Moreover, to further validate the effectiveness and generalization of Da Yu, we also include COCO captioning (132K images) for evaluation.

\textbf{Model Settings:} For image captioning models, We include models with various paradigms to evaluate the performances. 1. Normal captioning models: DLCT \cite{luo2021dual} and COS-Net \cite{li2022comprehending}; 2. Normal pretrained vision-language models: Oscar \cite{li2020oscar}, BLIP \cite{li2022blip} and BLIP-2 \cite{li2023blip}; 3. Multi-modal large language models: MobileVLM \cite{wu2024mobilevlm}, MobileVLM V2 \cite{chu2024mobilevlm2}, InternVL 2.5 (0.9B and 1.9B) \cite{chen2024internvl}, MiniCPM-V 2.0 (2.8B) \cite{yao2024minicpm}, Qwen-VL 2.5 (3B) \cite{bai2025qwen2} and our proposed Da Yu, including three versions (0.7B, 1.7B and 3.2B). For adaptor modules, we select the normal MLP, LDP \cite{wu2024mobilevlm}, LDPv2 \cite{chu2024mobilevlm2} and MWA \cite{long2023multiway} for the comparison.

\textbf{Training Settings:} For our proposed WaterCaption dataset, we resize the image as 448 $\times$ 448. We set the batch size as 32 per device and the  initial learning rate as 4e-5. We use the AdamW optimizer with the cosine scheduler for a total 10 epochs in the finetuning stage, where 9 epochs for the initial modality alignment phase and 1 epoch for the captioning adaptation phase. We set the warm-up ratio as 0.03. The training and test process are on four RTX 4090 GPUs. 

For the COCO captioning dataset, we train the models for 30 epochs with the learning rate between 1e-6 and 5e-7, companied with the cosine scheduler. We set the total batch size as 128.

\textbf{Evaluation Settings:} To comprehensively evaluate the performances of models, we include ROUGH (1, 2 and L) \cite{lin2004rouge}, BLEU (1-3) \cite{papineni2002bleu}, METEOR \cite{meteor}, CIDEr \cite{vedantam2015cider} and GPT-Score \cite{sima2024drivelm}, widely adopted in the text generation community.

\begin{table}
    \setlength\tabcolsep{7.0pt}
    \caption{Performances of Da Yu-L under different waterway situations.}
    \vspace{-3mm}
    \centering
    \begin{tabular}{c|cccc}
    \toprule
      \textbf{Waterway} & \textbf{ROUGH-L} & \textbf{BLEU-3} & \textbf{METROR} & \textbf{GPT-Score}  \\
      \hline
      Canal & 0.43 & 0.35 & 0.48 & 0.85 \\
      Moat & 0.32 & 0.24 & 0.41 & 0.79 \\
      River & 0.40 & 0.28 & 0.45 & 0.80\\
      Lake & 0.36 & 0.25 & 0.47 & 0.83\\
      \hline
      \textbf{Weather} & \textbf{ROUGH-L} & \textbf{BLEU-3} & \textbf{METROR} & \textbf{GPT-Score} \\
      \hline
      Sunny & 0.47 & 0.35 & 0.49 & 0.86 \\
      Rainy & 0.41 & 0.31 & 0.46 & 0.83 \\
      Overcast & 0.36 & 0.26 & 0.43 & 0.84 \\
      Snowy & 0.32 & 0.23 & 0.41 & 0.77 \\
      \hline
      \textbf{Lighting} & \textbf{ROUGH-L} & \textbf{BLEU-3} & \textbf{METROR} & \textbf{GPT-Score} \\
      \hline
      Normal & 0.46 & 0.32 & 0.49 & 0.85 \\
      Dim & 0.32 & 0.24 & 0.41 & 0.77 \\
      Strong & 0.40 & 0.30 & 0.45 & 0.81 \\
      \hline
      \textbf{Time-of-Day} & \textbf{ROUGH-L} & \textbf{BLEU-3} & \textbf{METROR} & \textbf{GPT-Score} \\
      \hline
      Daytime & 0.49 & 0.35 & 0.52 & 0.86 \\
      Nightfall & 0.36 & 0.27 & 0.43 & 0.81 \\
      Night & 0.33 & 0.22 & 0.39 & 0.74 \\
      \bottomrule
    \end{tabular}
    \label{tab:various_feature_compare}
\end{table}

\subsection{Quantitative Results}

\textbf{Overall performances on WaterCaption.} As Table \ref{tab:benchmark_compare} presents, we compare the performance of 15 different models on the WaterCaption dataset. Overall, we observe the following: First, conventional transformer-based models (such as DLCT and COS-Net) performed poorly, indicating their extremely limited capability in generating long-form textual descriptions. Second, although pretrained vision-language models like Oscar and BLIP demonstrated improved performance compared to the first category, they still lagged behind multi-modal large language models. This performance gap is evident in both syntactic structure metrics (e.g., ROUGE and BLEU) and semantic similarity metrics (e.g., CIDEr, METEOR, and GPT-Score). Third, we categorize the MLLMs into three groups based on their parameter scale. Overall, we find that models with approximately 3 billion parameters consistently produced higher-quality captions in terms of both syntactic structure and contextual semantics, outperforming those with around 1.* billion or fewer than 1 billion parameters. Our proposed Da Yu series models, across the S, B, and L variants, achieved consistently strong performance. Additionally, models such as InternVL, Qwen-VL, and MiniCPM-V also deliver competitive results.

\begin{table}
    \setlength\tabcolsep{7.0pt}
    \caption{Performances of different adaptors on Da Yu-L.}
    \vspace{-3mm}
    \centering
    \begin{tabular}{c|cccc}
    \toprule
      \textbf{Adaptors} & \textbf{ROUGH-L} & \textbf{BLEU-3} & \textbf{METROR} & \textbf{GPT-Score}  \\
      \hline
      MLP & 0.36 & \textbf{0.29} & 0.43 & 0.81 \\
      LDP & 0.32 & 0.24 & 0.42 & 0.78\\
      LDPv2 & 0.36 & 0.27 & 0.46 & 0.82 \\
      MWA & 0.32 & 0.26 & 0.46 & \textbf{0.83} \\
      \hline
      \textbf{NTA (ours)} & \textbf{0.38} & 0.28 & \textbf{0.46} & \textbf{0.83} \\
      \bottomrule
    \end{tabular}
    \label{tab:dayu_adaptor}
\end{table}

\textbf{Performances under different waterway scenarios.} Table \ref{tab:datasets_char_compare} shows the performances of Da Yu-L under various waterway situations. We observe that for relatively structured waterways, such as canals and rivers, the captioning task poses significantly less difficulty for the models. In contrast, for more irregular and unstructured environments like moats and lakes, the quality of generated captions was comparatively lower. Secondly, the models produce high-quality captions under clear conditions; however, performances decline under adverse weather scenarios such as rain, overcast skies, and snowfall. Similar degradation in caption quality is also observed under varying lighting conditions and different times of day.

\begin{table}
    \setlength\tabcolsep{7.0pt}
    \caption{Performances of portability for NTA on MobileVLM v2.}
    \vspace{-3mm}
    \centering
    \begin{tabular}{c|cccc}
    \toprule
      \textbf{Adaptors} & \textbf{ROUGH-L} & \textbf{BLEU-3} & \textbf{METROR} & \textbf{GPT-Score}  \\
      \hline
      LDP & 0.32 & 0.21 & 0.40 & 0.72 \\
      LDPv2 & \textbf{0.34} & 0.24 & 0.41 & 0.74\\
      \hline
      \textbf{NTA (ours)} & \textbf{0.34} & \textbf{0.26} & \textbf{0.43} & \textbf{0.78} \\
      \bottomrule
    \end{tabular}
    \label{tab:nta_on_mobilevlm}
\end{table}

\begin{table}
    \setlength\tabcolsep{10.0pt}
    \caption{Comparison of computational complexity for different adaptors. The size of $q$ after adaptive pooling from $Q$ is set to 12 $\times$ 12.}
    \vspace{-3mm}
    \centering
    \begin{tabular}{c|ccc}
    \toprule
      \textbf{Adaptors} & \textbf{Types} & \textbf{Params (M) $\downarrow$} &   \textbf{FLOPs (G) $\downarrow$}  \\
      \hline
      MLP & linear & 7.23 & 3.22\\
      LDP & convolution & 18.94 & 7.87\\
      LDPv2 & convolution & 6.32 & 4.25 \\
      MWA & attention & 16.79 & 8.48 \\
      \hline
      \textbf{NTA (ours)} & \textbf{hybrid} & \textbf{4.81} & \textbf{2.46} \\
      \bottomrule
    \end{tabular}
    \label{tab:dayu_adaptor}
\end{table}

\textbf{Performances of Nano Transformer Adaptor (NTA).} Table \ref{tab:dayu_adaptor} presents the performances of different adaptors in Da Yu-L. Our proposed NTA achieves the best overall performance, significantly enhancing the LLM's capabilities in both syntactic structure prediction and contextual semantic generation. Additionally, the MLP module based on linear layers and the MWA module based on self-attention also deliver competitive results, with performance on BLEU-3 and GPT-Score metrics closely approaching that of our approach. Subsequently, Table \ref{tab:nta_on_mobilevlm} presents the performance of our proposed NTA when integrated with MobileVLM v2. It is evident that NTA significantly outperforms the native LDP-series adaptors of MobileVLM, demonstrating its strong transferability across different model architectures. In terms of computational efficiency, as shown in Table \ref{tab:dayu_adaptor}, NTA contains only 4.81 million parameters, which is substantially fewer than other adaptors, regardless of whether they are based on linear layers, convolutional structures, or attention mechanisms. Moreover, NTA’s FLOPs are as low as 2.46G, significantly lower than those of other adaptors, indicating its capability for highly efficient inference.

\begin{table}
    \centering
    \setlength\tabcolsep{0.4pt}
    \caption{GPT-Score for different regions of waterway scenarios.}
    \vspace{-3mm}
    \begin{tabular}{c|c|c|c}
    \toprule
       \textbf{Regions}  & MobileVLM v2 1.4B & InternVL 2.5 1.9B & Da Yu-B 1.7B \\
    \hline
       Weather  &  0.87 & \textbf{0.92} & \textbf{0.92} \\
       Shore condition & 0.63 & 0.66 & \textbf{0.70} \\
       Waterway buildings & 0.81 & \textbf{0.83} & \textbf{0.83} \\
       Waterway objects & 0.83 & 0.87 & \textbf{0.90} \\
       Wave amplitude & 0.87 & \textbf{0.90} & \textbf{0.90} \\
       Waterway cleanliness & 0.71 & 0.75 & \textbf{0.78} \\
       Ego-USV position & 0.80 & 0.86 & \textbf{0.89} \\
     \bottomrule
    \end{tabular}
    \label{tab:gpt_score_regions}
\end{table}

\begin{table}
    \centering
    \setlength\tabcolsep{4.5pt}
    \caption{Comparison of different finetuning methods for Da Yu-L.}
    \vspace{-3mm}
    \begin{tabular}{c|cccc}
    \toprule
      \textbf{Finetuning Methods} & \textbf{ROUGH-L} & \textbf{BLEU-3} & \textbf{METROR} & \textbf{GPT-Score}  \\
    \hline
      LoRA    & 0.36 & \textbf{0.30} & 0.43 & 0.81 \\
      Adaptor & \textbf{0.38} & 0.28 & \textbf{0.46} & \textbf{0.83} \\
    \bottomrule
    \end{tabular}
    \label{tab:finetune_compare}
\end{table}

\begin{table}
    \centering
    \setlength\tabcolsep{11.1pt}
    \caption{Finetuning performances of models on COCO dataset.}
    \vspace{-3mm}
    \begin{tabular}{cc|ccc}
    \toprule
     \textbf{Models} & \textbf{Params} & \textbf{BLEU-4} & \textbf{CIDEr} & \textbf{SPICE} \\
     \hline
     CLIPCap & 156M & 0.34 & 1.13 & 0.21\\
     BLIP & 583M & 0.40 & 1.33 & 0.24 \\
     BLIP-2 & 1.1B & \textbf{0.45} & 1.40 & 0.30\\
     InternVL 2.5 & 1.9B & 0.43 & 1.38 & 0.32\\
     \hline
     Da Yu-B & 1.7B & 0.43 & \textbf{1.41} & \textbf{0.33} \\
     \bottomrule
    \end{tabular}
    \label{tab:coco_compare}
\end{table}

\textbf{Image captioning quality for different regions of waterway.} To comprehensively evaluate the caption generation quality for different regions of waterway, we uniformly sampled 500 instances from the test set across various waterway scenarios. For each instance, the captions corresponding to the seven described regions in both the generated output and the ground truth were separated and individually evaluated using GPT-4o. We found that Da Yu demonstrated relatively accurate understanding and description in four key aspects: weather conditions, waterborne objects, wave amplitude, and ego-USV position, with GPT-Scores consistently above 0.90. In contrast, captioning for shore condition and waterway cleanliness proved more challenging, achieving scores of only 0.72 and 0.80, respectively. Captions related to waterway buildings maintained a moderate level of quality. Overall, the three models exhibited a nearly identical distribution in their performance across different waterway regions, clearly highlighting the inherent difficulty and challenges presented by the WaterCaption dataset.

\textbf{Comparison of finetuning methods.} We compare the training results of the Nano Transformer Adaptor (NTA) with those obtained using LoRA. As shown in Table \ref{tab:finetune_compare}, adaptor-based training generally achieves superior performance, indicating that our proposed NTA effectively bridges the gap between visual and textual features.

\begin{table}
    \setlength\tabcolsep{5.5pt}
    \caption{Comparison of inference latency on NVIDIA Jetson Orin. The LLM of VLMs is quantized to 4-bit.}
    \vspace{-3mm}
    \centering
    \begin{tabular}{c|c|c|c}
    \toprule
    \textbf{Models} & InternVL 2.5 0.9B &  \textbf{Da Yu-S 0.7B} & -  \\
      \midrule
      \textbf{Time (s)} & 6.81 & \textbf{4.35} & - \\
      \midrule
      \textbf{Models} & LLaVA-1.5 1.4B & MobileVLM v2 1.4B &   \textbf{Da Yu-B 1.7B}  \\
      \midrule
      \textbf{Time (s)} & 11.67 & \textbf{5.32} & 6.10 \\
      \midrule
      \textbf{Models} & Qwen 2.5 3B & MiniCPM-V 2.8B &   \textbf{Da Yu-L 3.2B}  \\
      \midrule
      \textbf{Time (s)} & 24.31 & \textbf{12.27} & 16.58 \\
      \bottomrule
    \end{tabular}
    \label{tab:inference_speed}
\end{table}

\begin{figure*}
    \includegraphics[width=0.99\linewidth]{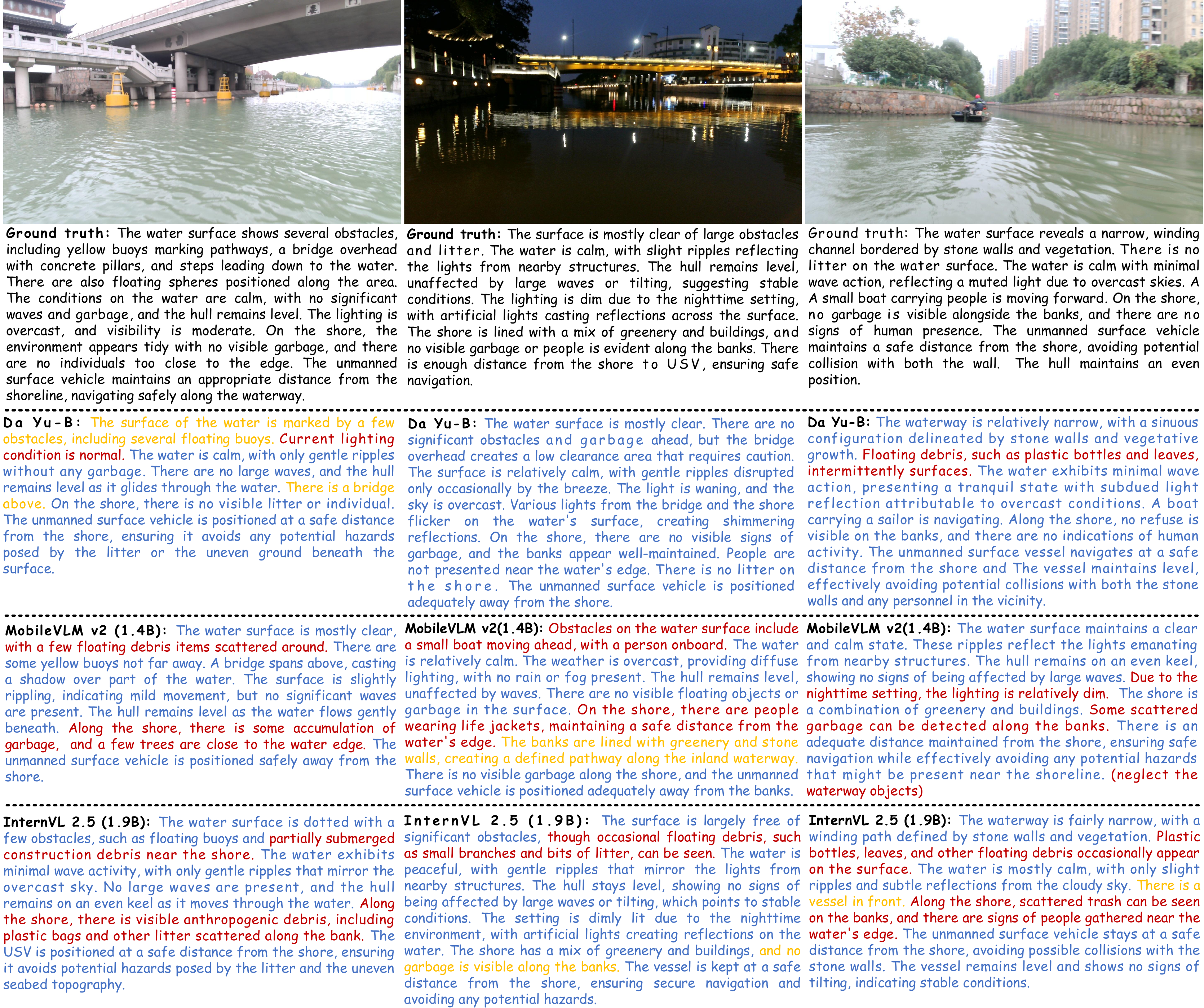}
    \vspace{-3mm}
    \caption{Prediction results of various multi-modal large language models under different waterway scenarios. \textcolor{blue}{Text segments highlighted in blue} indicate entirely correct captions. \textcolor{red}{Segments in red} denote captions that are completely counterfactual or contain significant omissions. \textcolor{orange}{Segments in orange} represent captions that are partially inaccurate or biased.}
    \label{fig:overall_experiments}
\end{figure*}

\textbf{Inference time of MLLMs.} As shown in Figure \ref{tab:inference_speed}, we compare the inference speed of various MLLMs on the WaterCaption dataset. For models with fewer than 1 billion parameters, our Da Yu-S achieves a latency of 4.35 seconds, outperforming InternVL 2.5 (0.9B). Da Yu-B records a latency of 6.10 seconds, slightly higher than MobileVLM v2 (1.4B) at 5.32 seconds, but significantly faster than LLaVA-1.5 (1.4B). Da Yu-L exhibited slightly slower inference than MiniCPM-V (2.8B), yet faster than Qwen 2.5 (3B). Overall, the Da Yu models demonstrate the capability to support real-time, fine-grained monitoring log generation for unmanned surface vehicles operating in inland waterways.

\textbf{Generalization performances of Da Yu.} As shown in Figure \ref{tab:coco_compare}, we evaluate the model's performance on the COCO Captioning dataset, which includes both models specifically designed for short-text captioning and general-purpose MLLMs such as InternVL 2.5. Da Yu also achieves strong performance in this setting, demonstrating its robust generalization capability beyond the WaterCaption domain.

\subsection{Ablation Studies}

To further validate the effectiveness of the Da Yu model and framework, we present the results of the ablation study in Table \ref{tab:ablation_experiments}.

\begin{table}
\setlength\tabcolsep{8.5pt}
    \label{tab:ablation_experiments}
    \centering
    \caption{Ablation experiments based on Da Yu-L.}
    \vspace{-3mm}
    \begin{tabular}{c|ccc}
    \toprule
      \textbf{Methods} & \textbf{BLEU-3} &\textbf{METROR} & \textbf{GPT-Score} \\
    \hline
        Baseline & \textbf{0.28} & \textbf{0.46} & \textbf{0.83} \\
    \hline
      \multicolumn{4}{c}{\textbf{NTA}}   \\
    \hline
      No GDC & \textbf{0.30} & 0.45 & 0.81 \\
      Size of $q$ = 8 $\times$ 8 & 0.25 & 0.45 & 0.80 \\
      PA $\rightarrow$ Vanilla CA & 0.26 & 0.42 & \textbf{0.84} \\
    \hline
      \multicolumn{4}{c}{\textbf{Visual Feature Encoding}} \\
    \hline
      No pixel shuffle & 0.25 & 0.44 & \textbf{0.83} \\
      AVE $\rightarrow$ Normal patch & 0.22 & 0.43 & 0.79 \\
    \hline
      \multicolumn{4}{c}{\textbf{Training Process}} \\
    \hline
      IMAP + CAP $\rightarrow$ CAP & 0.27 & 0.44 & 0.82 \\
    \bottomrule
    \end{tabular}

\end{table}

For the Nano Transformer Adaptor (NTA), first, the removal of the grouped dilated convolution (GDC) leads to a noticeable decline in both METEOR and GPT-Score, indicating its importance in enhancing semantic representation. Second, when reducing the size of the query $q$ from the baseline $12 \times 12$ to $8 \times 8$, all three evaluation metrics show a decline. This can be attributed to the reduced resolution of $q$, which limits the amount of informative content available for softmax attention computation with $K$ and $V$, thereby restricting the model’s ability to retrieve relevant features and degrading overall representational capacity. Third, when replacing the pooling attention mechanism with vanilla cross-attention, although GPT-Score slightly improved, the other two metrics declined, suggesting that pooling attention provides a better balance between efficiency and effectiveness in this context.

Regarding visual feature encoding, when the pixel shuffle operation is removed, we observe a decrease in both BLEU-3 and METEOR scores. Moreover, when the adaptive visual encoding was replaced with standard visual patch extraction, all three evaluation metrics experience a significant decline, highlighting the effectiveness of the proposed adaptive encoding strategy in enhancing captioning performance.

With respect to the training process, when we replace the two-stage strategy—comprising the Initial Modality Alignment Phase and the Captioning Adaptation Phase, with a single-stage approach that begins directly with the Captioning Adaptation Phase, all three evaluation metrics shows a noticeable decline. This highlights the critical role of the initial modality alignment phase and underscores the effectiveness and superiority of our proposed two-stage training strategy.

\begin{figure}
    \includegraphics[width=0.99\linewidth]{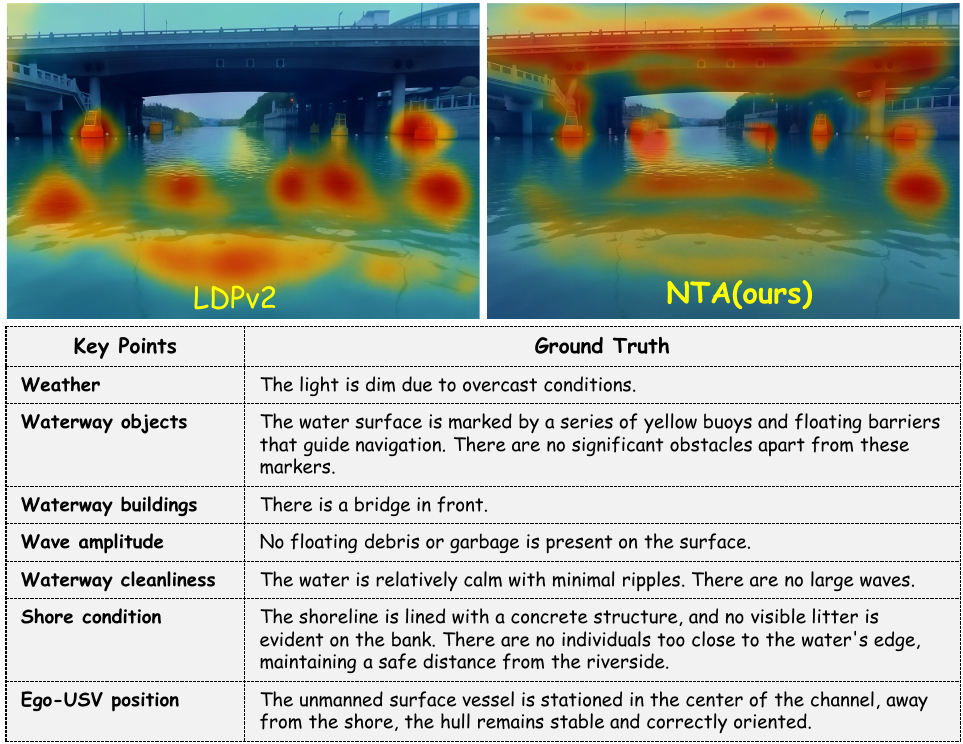}
    \vspace{-3mm}
    \caption{The visualization of heatmap in two different adaptor modules, including LDPv2 and our proposed NTA.}
    \label{fig:heatmap}
\end{figure}

\begin{figure}
    \includegraphics[width=0.99\linewidth]{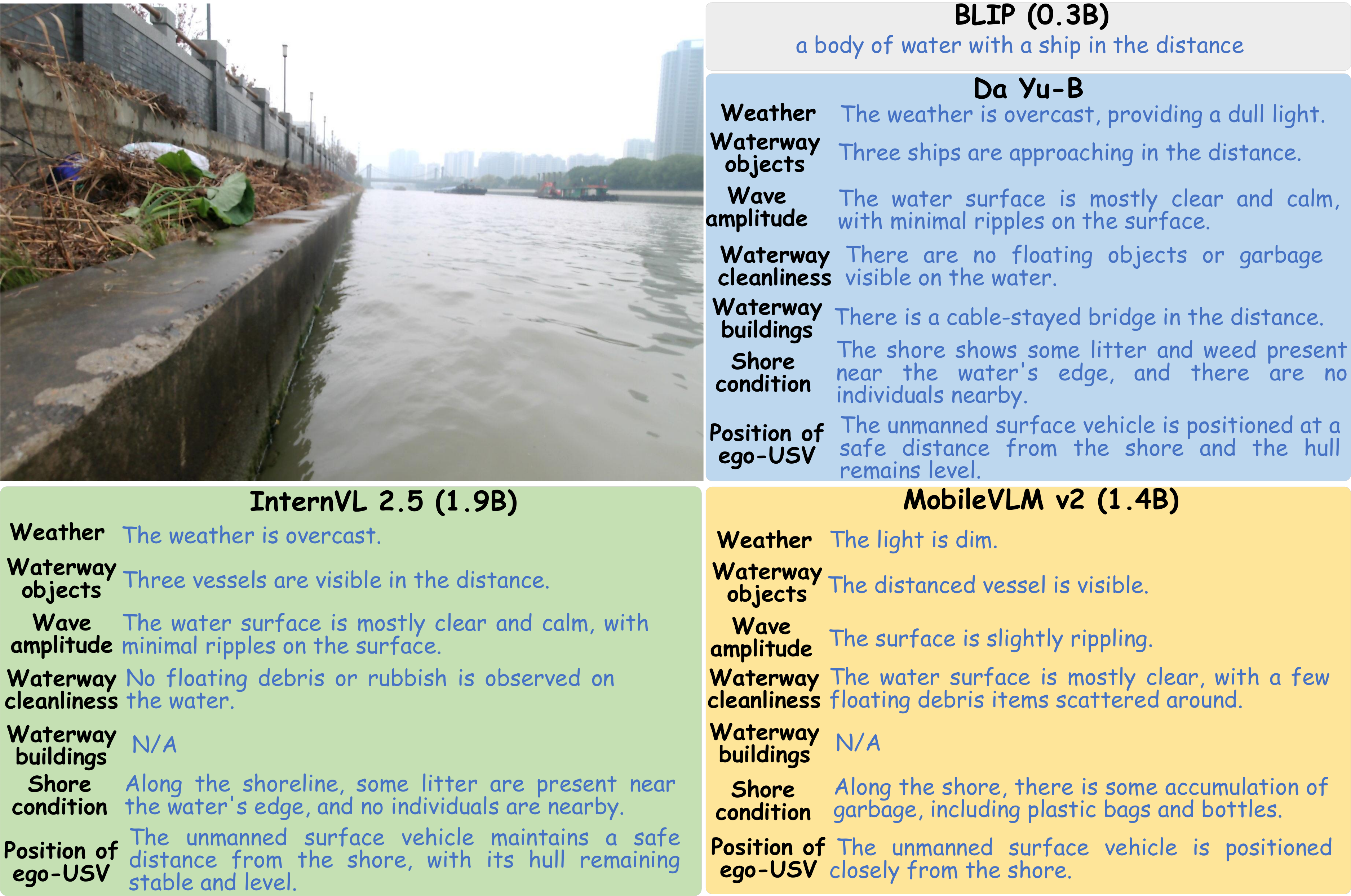}
    \vspace{-3mm}
    \caption{Comparison between different models on waterway image captioning for different key elements.}
    \label{fig:exact_one_sample}
\end{figure}

\subsection{Visualization Results}
As shown in Fig. \ref{fig:overall_experiments}, we compare the captioning performance of different MLLMs across diverse waterway environments. Overall, the proposed Da Yu model demonstrates the best capability in generating fine-grained captions. Although a few counterfactual and partially biased descriptions still occur, Da Yu consistently outperforms both MobileVLM v2 and InternVL, exhibiting the best robustness among the compared models. In contrast, MobileVLM v2 and InternVL tend to suffer from more hallucinations, resulting in inaccurate or counterfactual outputs. For instance, in the third sample, InternVL incorrectly identifies riverside elements, misclassifying certain facilities as people and falsely detecting floating debris on a water surface that is actually clean.

Fig. \ref{fig:heatmap} presents a comparison of heatmaps generated by Da Yu-B using two different adaptors, LDPv2 and NTA, during inference on the same image. We observe that NTA effectively captures various relevant and accurate locations within the waterway. Specifically, Da Yu-B equipped with LDPv2 misses several buoys located in the front-left region and fails to attend to the bridge ahead. In contrast, the version with NTA focuses on more reasonable and relevant target areas.

Fig. \ref{fig:exact_one_sample} presents a category-wise analysis of the captions generated by different models across various regions of a waterway scene. We observe that BLIP focuses solely on the vessels in the water, completely neglecting other areas. Among the three MLLMs (Da Yu-B, InternVL 2.5, and MobileVLM v2), all models provide consistent and accurate descriptions of the weather. In terms of waterway objects, Da Yu-B offers more precise descriptions, including the types and orientations of the vessels. For wave amplitude, waterway cleanliness, and the position of the ego-USV, the captions generated by all three models are generally consistent. However, only Da Yu-B identifies the distant bridge under the category of waterway buildings. Regarding the shoreline, all three models detect the presence of litter, but MobileVLM v2 stands out by successfully specifying the type of garbage.

\begin{figure}
    \includegraphics[width=0.99\linewidth]{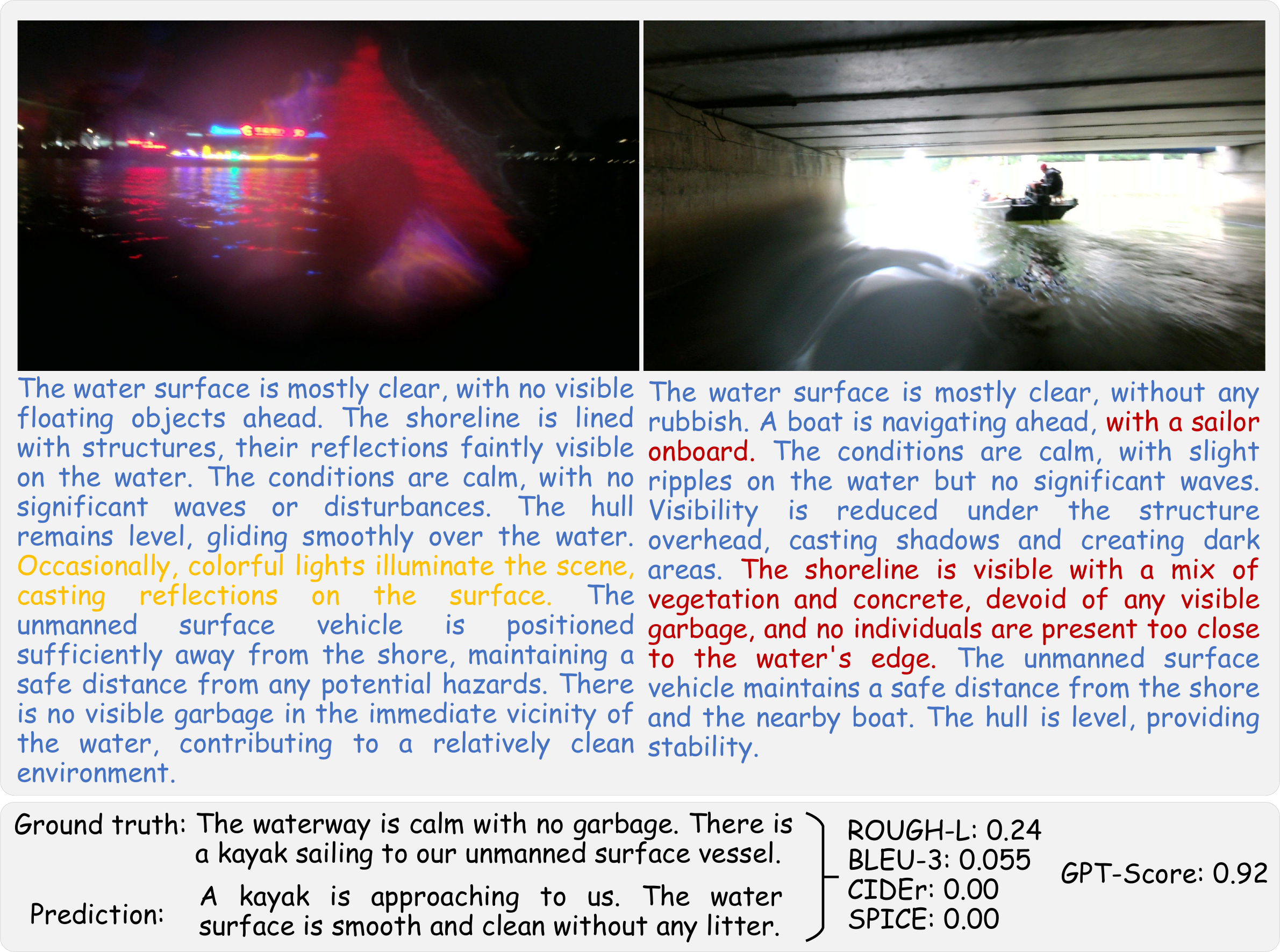}
    \vspace{-3mm}
    \caption{The limitations in WaterCaption dataset. The text is the predicted caption by Da Yu-B. \textcolor{blue}{Text segments highlighted in blue} indicate entirely correct captions. \textcolor{red}{Segments in red} denote captions that are completely counterfactual or contain significant omissions. \textcolor{orange}{Segments in orange} represent captions that are partially inaccurate or biased.}
    \label{fig:limitations}
\end{figure}

\section{Limitations and Future Works}
\label{sec:limitations}

\subsection{Limitations}

The limitations of our approach can be categorized into three main aspects.

First, as shown in Figure \ref{fig:limitations}, the model tends to hallucinate and produce counterfactual captions under challenging conditions such as camera blur, shake, darkness, or glare. For example, in the left image, color distortion caused by water mist on the lens is mistakenly interpreted as part of the actual scene. In the right image, due to the vessel being under a bridge in a dark area, the camera fails to detect the front vessel as it emerges from the shaded region into a brightly lit area, resulting in biased captions due to glare-induced blind spots.

Second, since the work focuses on generating fine-grained and long-form captions, the model struggles to produce accurate captions for out-of-distribution scenes. This task is dramatically tougher than generating short-form image captions.

Third, as shown in all the quantitative tables earlier, the GPT-Score metric outperforms conventional evaluation metrics that rely on textual window matching or word-level similarity (e.g., ROUGE, BLEU) when assessing long-form captions. Traditional metrics often fail to capture the overall semantic correctness of lengthy descriptions, while GPT-Score leverages the strong reasoning capabilities of large language models to make more accurate judgments. For example, when a model generates a scene with incorrect temporal or spatial order—as illustrated in the bottom sample of Figure 13, conventional metrics yield very low scores, whereas GPT-Score aligns closely with human judgment.
However, the reliability of large language models in evaluation remains an open question. Therefore, we argue that it is essential to design a dedicated, reliable, and interpretable evaluation metric specifically tailored for long-form image captioning tasks.

\section{Conclusion}
\label{sec:conclusion}
In this work, we introduce the first image captioning dataset specifically designed for automated waterway monitoring based on unmanned surface vehicles (USVs). This dataset is also the first to support fine-grained, multi-region captioning. It presents a unique challenge: generating detailed and accurate monitoring reports based on waterway images under predefined specifications.
Furthermore, we propose Da Yu, the first multi-modal large language model (MLLM) tailored for the domain of waterborne traffic perception. Da Yu is capable of efficient inference on edge devices and achieves state-of-the-art performance on the proposed image captioning dataset.
Within the Da Yu model, we introduce a novel adaptor module named Nano Transformer Adaptor (NTA), which effectively extracts and fuses both global and local features. NTA enables low-computation alignment between image and text modalities while maintaining strong accuracy and generalization capabilities.
In future work, we plan to extend this framework toward developing a general-purpose multimodal large language model for multi-task perception in aquatic traffic environments.

% if have a single appendix:
%\appendix[Proof of the Zonklar Equations]
% or
%\appendix  % for no appendix heading
% do not use \section anymore after \appendix, only \section*
% is possibly needed

% use appendices with more than one appendix
% then use \section to start each appendix
% you must declare a \section before using any
% \subsection or using \label (\appendices by itself
% starts a section numbered zero.)
%

\appendices

% use section* for acknowledgment
% \section*{Acknowledgment}

% The authors would like to thank...

% Can use something like this to put references on a page
% by themselves when using endfloat and the captionsoff option.
\ifCLASSOPTIONcaptionsoff
  \newpage
\fi

% trigger a \newpage just before the given reference
% number - used to balance the columns on the last page
% adjust value as needed - may need to be readjusted if
% the document is modified later
%\IEEEtriggeratref{8}
% The "triggered" command can be changed if desired:
%\IEEEtriggercmd{\enlargethispage{-5in}}

% references section

% can use a bibliography generated by BibTeX as a .bbl file
% BibTeX documentation can be easily obtained at:
% http://mirror.ctan.org/biblio/bibtex/contrib/doc/
% The IEEEtran BibTeX style support page is at:
% http://www.michaelshell.org/tex/ieeetran/bibtex/
\normalem
\footnotesize
\bibliographystyle{IEEEtran}
% argument is your BibTeX string definitions and bibliography database(s)
\bibliography{refs}
%
% <OR> manually copy in the resultant .bbl file
% set second argument of \begin to the number of references
% (used to reserve space for the reference number labels box)
\vspace{-12mm}

\begin{IEEEbiographynophoto}{Runwei Guan (Member, IEEE)} is currently a research fellow affiliated at Thrust of AI, HKUST(GZ). He received his PhD degree from University of Liverpool in 2024 and M.S. degree in Data Science from University of Southampton in 2021. His research interests include radar perception, multi-sensor fusion, vision-language learning, lightweight neural network, and statistical machine learning. He serves as the peer reviewer of TITS, TNNLS, TIV, TCSVT, ITSC, ICRA, RAS, EAAI, MM, etc.
\end{IEEEbiographynophoto}
\vspace{-12mm}

\begin{IEEEbiographynophoto}{Ningwei Ouyang} is pursuing the Ph.D. degree in the University of Liverpool. His research interest includes computer vision and pattern recognition.
\end{IEEEbiographynophoto}
\vspace{-12mm}

\begin{IEEEbiographynophoto}{Tianhao Xu} is currently a research assistant of Hong Kong University of Science and Technology (GuangZhou).
\end{IEEEbiographynophoto}
\vspace{-12mm}

\begin{IEEEbiographynophoto}{Shaofeng Liang} is currently at the Qingdao Institute of Software, China University of Petroleum (East China) pursuing a master's degree. His research interests include multi-modal object tracking.
\end{IEEEbiographynophoto}
\vspace{-12mm}

\begin{IEEEbiographynophoto}{Wei Dai} is currently working toward the Ph.D. degree with the School of Advanced Technology, XJTLU. 
\end{IEEEbiographynophoto}
\vspace{-12mm}

\begin{IEEEbiographynophoto}{Yafeng Sun} is pursuing the M.S. degree in Cyberspace Security at the University of Science and Technology of China.
\end{IEEEbiographynophoto}
\vspace{-12mm}

\begin{IEEEbiographynophoto}{Shang Gao} is a Research Assistant at The Hong Kong University of Science and Technology (Guangzhou). 
\end{IEEEbiographynophoto}
\vspace{-12mm}

\begin{IEEEbiographynophoto}{Songning Lai} is currently pursuing a Ph.D degree at the School of AI Thrust, HKUST(GZ), China. 
\end{IEEEbiographynophoto}
\vspace{-12mm}

\begin{IEEEbiographynophoto}{Shanliang Yao} is currently a lecturer with the school of information engineering, Yancheng Institute Technology, Yancheng, China. 
\end{IEEEbiographynophoto}
\vspace{-12mm}

\begin{IEEEbiographynophoto}{Xuming Hu} received the Ph.D. degree from the Tsinghua University, in 2024. He is currently an Assistant Professor with the Thrust of Artificial Intelligence, HKUST(GZ).
\end{IEEEbiographynophoto}
\vspace{-12mm}

\begin{IEEEbiographynophoto}{Ryan Wen Liu} the Ph.D. degree from The Chinese University of Hong Kong, Hong Kong, in 2015. He is currently a Professor with the School of Navigation, Wuhan University of Technology. 
\end{IEEEbiographynophoto}
\vspace{-12mm}

\begin{IEEEbiographynophoto}{Yutao Yue} (Senior Member, IEEE) is an associate professor at the Artificial Intelligence Thrust and Intelligent Transportation Thrust of HKUST(GZ).
\end{IEEEbiographynophoto}
\vspace{-12mm}

\begin{IEEEbiographynophoto}{Hui Xiong} (Fellow, IEEE) is currently serves as the Associate Vice-President for Knowledge Transfer at the Hong Kong University of Science and Technology (Guangzhou), Chair Professor of the Artificial Intelligence Thrust, Director of the AI+ Lab, and the Founding Head of the Artificial Intelligence Thrust. Previously, he was a distinguished professor for tenure at Rutgers University. He has published over 400 papers in data science and artificial intelligence, with over 53,000 Google Citations and an h-index of 96. His accolades include Fellow of the AAAI, AAAS, IEEE, and CAAI, the Distinguished Scientist of the Association for Computing Machinery (ACM), and received highly recognized awards such as the 2023 ACM SIGKDD Service Award, the 2021 AAAI Best Paper Award and IEEE ICDM Best Paper Award (2011) for his contributions to data mining and mobile computing. Dr. Xiong currently serves as the ACM SIGKDD Secretary, Co-Editor-in-Chief of the Encyclopedia of GIS, and Editor-in-Chief of Nature npj | Artificial Intelligence. He has held significant roles in organizing major conferences, including as Program Co-Chair for the ACM SIGKDD and IEEE ICDM. Moreover, he served as the Deputy Dean at Baidu Research, where he played a crucial role in guiding the operations and strategic direction of five research laboratories. Additionally, he has supervised nearly 30 Ph.D. students, many of whom have secured faculty positions at prestigious universities worldwide, including the University of Arizona, Stony Brook University, and the University of Tennessee – Knoxville.
\end{IEEEbiographynophoto}

% biography section
% 
% If you have an EPS/PDF photo (graphicx package needed) extra braces are
% needed around the contents of the optional argument to biography to prevent
% the LaTeX parser from getting confused when it sees the complicated
% \includegraphics command within an optional argument. (You could create
% your own custom macro containing the \includegraphics command to make things
% simpler here.)
%\begin{IEEEbiography}[{\includegraphics[width=1in,height=1.25in,clip,keepaspectratio]{mshell}}]{Michael Shell}
% or if you just want to reserve a space for a photo:

% \begin{IEEEbiographynophoto}{Runwei Guan (Member, IEEE)} is currently a research fellow affiliated at Thrust of AI, Hong Kong University of Science and Technology (GuangZhou). He received his PhD degree from University of Liverpool in 2024. His research interests include multi-sensor perception, vision-language learning, multi-task learning. He serves as the peer reviewer of TITS, TNNLS, TIV, TCSVT, ITSC, ICRA, RAS, EAAI, MM, etc.
% \end{IEEEbiographynophoto}

% You can push biographies down or up by placing
% a \vfill before or after them. The appropriate
% use of \vfill depends on what kind of text is
% on the last page and whether or not the columns
% are being equalized.

%\vfill

% Can be used to pull up biographies so that the bottom of the last one
% is flush with the other column.
%\enlargethispage{-5in}

% that's all folks
\end{document}